\documentclass{sn-jnl}% Math and Physical Sciences Reference Style
\usepackage{graphicx}%
\usepackage{multirow}%
\usepackage{amsmath,amssymb,amsfonts}%
\usepackage{amsthm}%
\usepackage{mathrsfs}%
\usepackage[title]{appendix}%
\usepackage{soul}%
\usepackage{xcolor}%
\usepackage{textcomp}%
\usepackage{manyfoot}%
\usepackage{booktabs}%
\usepackage{algorithm}%
\usepackage{algorithmicx}%
\usepackage{algpseudocode}%
\usepackage{listings}%
\usepackage{array}
\usepackage{natbib}
\usepackage[most]{tcolorbox}
\tcbuselibrary{breakable}

\usepackage[justification=centering]{caption}
\theoremstyle{thmstyleone}%
\theoremstyle{thmstyletwo}%

\begin{document}

\title[Article Title]{Elementary Math Word Problem Generation using Large Language Models}


% \author*[1,2]{\fnm{Nimesh} \sur{Ariyarathne}}\email{ariyarathnanimesh@gmail.com}

\author[1]{\fnm{*Nimesh} \sur{Ariyarathne}}\email{nimeshariyarathne.19@cse.mrt.ac.lk}

\author[1]{\fnm{*Harshani} \sur{Bandara}}\email{harshani.19@cse.mrt.ac.lk}

\author[1]{\fnm{*Yasith} \sur{Heshan}}\email{yasith.19@cse.mrt.ac.lk}

\author[2]{\fnm{*Omega} \sur{Gamage}}\email{omega.gamage@accelr.site}

\author[1]{\fnm{Dilan} \sur{Nayanajith}}

\author[1]{\fnm{Yutharsan} \sur{Sivapalan}}

\author[1]{\fnm{Gayathri} \sur{Lihinikaduarachchi}}

\author[1]{\fnm{Tharoosha} \sur{Vihidun}}

\author[1]{\fnm{Meenambika} \sur{Chandirakumar}}
 
\author[1]{\fnm{Sanujen} \sur{Premakumar}} 

\author[1]{\fnm{Sanjula} {Gathsara}} 

\author[3]{\fnm{*Surangika} \sur{Ranathunga}}\email{s.ranathunga@massey.ac.nz}

\affil*[1]{\orgdiv{Department of Computer Science and Engineering}, \orgname{University of Moratuwa}, \orgaddress{ \city{Katubedda}, \postcode{10400}, \country{Sri Lanka}}}

\affil*[2]{\orgdiv{Acceler Logic}, \city{Colombo}, \country{Sri Lanka}}

\affil*[3]{\orgdiv{School of Mathematical and Computational Sciences}, \orgname{Massey University}, \orgaddress{ \city{Auckland}, \country{New Zealand}}}

\affil[]{\footnotemark[1] Core authors}

\newpage
% previous code
\abstract{Mathematics is often perceived as a complex subject by students, leading to high failure rates in exams. To improve Mathematics skills, it is important to provide sample questions for students to practice problem-solving. Manually creating Math Word Problems (MWPs) is time consuming for tutors, because they have to type in natural language while adhering to grammar and spelling rules of the language. Early techniques that use pre-trained Language Models for MWP generation either require a tutor to provide the initial portion of the MWP, and/or additional information such as an equation. In this paper, we present an MWP generation system (MathWiz) based on Large Language Models (LLMs) that overcomes the need for additional input - the only input to our system is the number of MWPs needed, the grade and the type of question (e.g.~addition, subtraction). Unlike the existing LLM-based solutions for MWP generation, we carried out an extensive set of experiments involving different LLMs, prompting strategies, techniques to improve the diversity of MWPs, as well as techniques that employ human feedback to improve LLM performance. Human and automated evaluations confirmed that the generated MWPs are high in quality, with minimal spelling and grammar issues. However, LLMs still struggle to generate questions that adhere to the specified grade and question type requirements. }

\maketitle
\section{Introduction}\label{sec1}
{Mathematics is often perceived as a complex subject by students, leading to high
failure rates in exams ~\citep{casinillo2019factors}. To improve Mathematics skills, it is important to provide
sample questions for students to practice problem-solving. This requires tutors to come up with a diverse set of Mathematics questions manually or by using automated mechanisms.

Math Word Problems (MWPs), as the ones shown in Figure~\ref{fig:mwp_example}, are an important type of educational resource that helps assess and improve
students’ proficiency in various mathematical concepts and skills \citep{wang2021math}. An MWP is a mathematical problem expressed in natural language and it requires problem-solving, as well as language comprehension ability. However, given that an MWP contains a higher portion of natural language text, creating customized MWPs can be a time-consuming task  for the tutors. 

A viable alternative is to automatically generate MWPs. Pre-trained Language Models have shown promising results in this front. Early such systems require the user to provide the equation, starting portion (seed) of the MWP, or some context~\citep{wang2021math, niyarepola2022math}.  However, these methods entail limitations, particularly in the necessity to accurately define both a seed, equation and/or context that suits the MWP, which can be time-consuming. Furthermore, such methods may not guarantee the diversity of MWPs, nor they could cater to the requirement of generating MWPs while adhering to a particular grade or a question type.

In this context, generative AI techniques backed by Large Language Models (LLMs) provide a promising avenue. LLMs have created a paradigm shift in the field of Generative AI, with successful applications in many domains~\citep{gozalo2023survey}. However, for MWP generation, we are only aware of the work of~\citet{christ2024mathwell,hwang2024using, xie2024adversarial}.~\cite{xie2024adversarial} used LLMs to modify the values of a given MWP to derive a new MWP, but did not generate MWPs from scratch.~\citet{hwang2024using} simply prompted ChatGPT to generate MWPs, but did not fine-tune ChatGPT further.~\citet{christ2024mathwell} fine-tuned Llama-2 for MWP generation. Their model was able to generate questions at appropriate reading levels, but could not consider the question type. Their error analysis only considered three error types. Moreover, their model did not guarantee the diversity of the generated MWPs, nor did they go beyond basic fine-tuning.

{We present an MWP generation system, termed \textit{\textbf{MathWiz}}, which leverages the advanced text generation capabilities of LLMs. {Unlike the sequence-sequence Language Models used in early work, LLMs do not need the starting portion or the equation of the MWP to initiate the MWP generation process. Given the vast amount of knowledge they have accumulated through the pre-training process, they are capable of responding to a textual \textit{instruction} (written in the form of a \textit{prompt}) and carry out the task specified in the instruction. We exploit this capability of LLMs and } define well-curated textual prompts that clearly instruct the LLM to generate MWPs that adhere to a user-specified grade and a question type. In addition, we experiment with in-context learning, where the LLM is shown some sample MWPs, to better guide the LLM's text generation capabilities towards the user's requirements. } We carried out an extensive study to identify the best open-source LLM and prompt template for MWP generation. The most promising LLM was further fine-tuned with an MWP dataset that we manually created. We further improved the quality of the generated MWPs by incorporating human feedback to the training process. We also investigated on the hyper-parameters related to MWP generation, and determined the optimal parameter combination to guarantee the diversity of the generated questions, and incorporated a secondary LLM to guarantee the generated MWPs are solvable. 

\begin{figure}
    \centering
    \fbox{\begin{minipage}[t][1.7cm]{0,9\textwidth}
There are 6 red socks and 2 blue socks. How many socks are there?(Grade 1, Single-digit Addition)

Liam had 40 marbles, lost 15, then won 20 in a game. How many marbles does Liam have now? (Grade2, 2 - Digit Addition \& Subtraction)
\end{minipage}}
    \caption{Examples of MWPs (According to the United States Common Core Mathematics Curriculum)}
    \label{fig:mwp_example}
\end{figure}

In order to evaluate MathWiz, we carried out an extensive evaluation involving humans and an LLM-based evaluation technique, where the MWPs generated by the LLM after each improvement was evaluated. Human evaluation was conducted following the participatory research approach\footnote{The participants to the research becomes co-authors of the publication. This is increasingly becoming a common approach when a research does not have funds to compensate human participants~\cite{nekoto2020participatory}.}. Our evaluation considered 12 error categories, as opposed to the three error categories considered by~\citet{christ2024mathwell}. The human + LLM evaluation produced a dataset of 4K MWPs with error annotation. To the best of our knowledge, this is the first comprehensively error annotated English MWP corpus. This corpus, as well as the training MWP corpus that we manually created are publicly released under the \href{https://huggingface.co/datasets/MathWizards/MathWizard-mathword-problem-dataset-with-grade-section} {MathWizard dataset}.

%Mwp generation is done using different approaches  like  using the context and equation [https://arxiv.org/pdf/2109.04546.pdf], seed and
%In this paper, we are focusing on generating customized math word problems using the large language models LLMs. mainly we are focusing on math word problem generation for different grades according to their syllabus. For example, addition, subtraction, count, measurement and time are the types of MWPs for grade 1. Mwp related to the grade 1 addition question is given below.

%There are many types of research regarding MWP generation using the seed-based [omega’s paper], template-based [] question rewriting with different technologies  like
%This research is focused on creating MWPs in various domains as the solution for this time-consuming task. We aim to contribute to the broader goal of enhancing math education access and quality education for individuals in low-resource language communities. Addressing the unique challenges of math word problem generation using autoregressive [3] text generation techniques with Large language models (LLMs). we aim to overcome the limitations of the previous research [4] and contribute to advancing automatic Mathematics problem-generation systems
}
\section{Background and Related Work}\label{sec2}
In this section, we discuss the past work related to MWP generation. We also introduce the techniques and concepts related to LLM-based text generation, which were used in this research.

\subsection{MWP Generation}
Early approaches for MWP generation include question re-writing, template-based generation and generation with Neural Networks (NNs)~\citep{niyarepola2022math}. Early NN approaches include  training Deep Learning models such as Recurrent Neural Networks (RNNs) \citep{medsker2001recurrent, liyanage2020multi}, building custom NN models~\citep{qin2023mathematical}, and fine-tuning pre-trained Language Models such as mT5 \citep{xue2020mt5} or GPT~\citep{liu2023gpt}  for the task of MWP generation. In order to generate an MWP from such pre-trained Language Models, some form of input has to be provided to the model. This input can be in the form of an equation, context and/or a seed~\citep{wang2021math,omegas,niyarepola2022math, zhou2023learning}. However, these generated questions have general errors such as grammatical and spelling errors, as well as Mathematics-specific errors such as having wrong units, unsolvable questions and resulting in unrealistic answers~\citep{omegas}. Moreover, none of these NN approaches have the ability to generate MWPs to satisfy user requirements - such as an MWP belonging to a particular grade or a question type.
{
\subsection{LLMs for Text Generation}\label{llms_for_text_generation}

The two broad approaches for LLM-based text generation are auto-regressive and non-autoregressive text generation~\citep{xiao2023survey}. Auto-regressive text generation has shown to be more accurate and able to capture the sequential dependency relations among tokens, despite its higher latency during generation~\citep{chen2022harnessing}. Thus, auto-regressive text generation, which is employed by the modern-day LLMs, is the commonly used approach.

As of now, some of the LLMs that are considered state-of-the-art (SOTA) are{DeepSeek}~\citep{deepseekai2024deepseekllmscalingopensource}, Llama 1-4~\citep{touvron2023llama}, ~\citep{touvron2023llama2}), Mistral~\citep{jiang2023mistral}, Gopher~\citep{rae2021scaling}, Falcon~\citep{almazrouei2023falcon}, MPT, GPT-3~\citep{kalyan2023survey}, ChatGPT~\citep{kim2023chatgpt}, Chinchilla-70B~\citep{hoffmann2022training}, and PaLM-540B~\citep{chowdhery2023palm}. The effectiveness of these LLMs for text generation tasks has been compared in several research~\citep{hoffmann2022training, touvron2023llama, touvron2023llama2}. However, the best-performing LLM depends on the size of the LLM and the end task, among other factors.~ %\citet{touvron2023llama2} show that the accuracy in many evaluation matrices increases with the size of the Llama-2 model. Therefore, the Llama-2 70B model outperforms other Llama-2 models. Furthermore, this paper contains a comparison between the performance of the open source base models by evaluating using many evaluation matrices. The results show that the Llama-2 model is capable of providing more accurate results in most of the evaluation matrices by outperforming open-source models such as MPT, Falcon, and Llama 1.

As for generating MWPs with LLMs, we are only aware of the work of~\citet{christ2024mathwell,hwang2024using}. ~\citet{hwang2024using} used ChatGPT to generate MWPs belonging to three different difficulty levels. They designed a prompt that incorporates some contextual information. However, they did not fine-tune ChatGPT.~\citet{christ2024mathwell} fine-tuned Llama-2.0 (70B) for MWP generation. However, they did not go beyond basic fine-tuning. They evaluated the generated MWPs based on three key criteria:
\textit{Solvability}: Whether the problem can be solved using the information provided, ensuring it is clear, logical, and has a valid solution.
\textit{Accuracy}: Whether the solution and answers generated for the problem are correct and align with the expected outcome.
\textit{Appropriateness}: Whether the problem is relevant, meaningful, and suitable for the intended audience or context, avoiding ambiguity or confusion.

\subsection{Working with LLMs}
\label{sec:prompt_eng_lit}

%\subsection{LLM fine-tuning}
LLMs are built by training a NN (commonly used architecture is the Transformer~\citep{vaswani2017attention}) with large amounts of raw text. This step is commonly known as `pre-training'. Such LLMs do not have any task-specific ability. Therefore, this pre-trained LLM is further trained with task-specific data. This step is known as `fine-tuning'. A common fine-tuning technique is instruction tuning~\citep{chung2022scaling}, which involves fine-tuning an LLM on a collection of tasks described in the form of natural language instructions (written in the form of \textit{prompts}). Fine-tuning a full LLM (vanilla fine-tuning) is rather costly, due to the large parameter count of the modern-day LLMs.

As an alternative, Parameter Efficient fine-tuning (PEFT) techniques have been introduced~\citep{ding2023parameter}. Popular PEFT methods include Low Ranked Adapters (LoRA)~\citep{hu2021lora}, Prefix Tuning~\citep{liu2021p}, Prompt Tuning~\citep{lester2021power}, Bitfit~\citep{zaken2021bitfit}, \((IA)^3\)~\citep{liu2022few} and P-Tuning~\citep{liu2023gpt}. Out of these PEFT techniques,  LoRA~\citep{hu2021lora} has been quite popular recently. LoRA is an efficient adaptation strategy that allows for quick task-switching without introducing inference latency or reducing input sequence length. The proposed approach is applicable to any neural network with dense layers, thus lowering the hardware barrier. It allows for efficient task-switching and can be combined with other techniques. 
QLoRA (Quantized Low-Rank Adaptation) is a PEFT technique that further reduces computational and memory overhead~\citep{dettmers2023qlora}. It combines quantization and low-rank adaptation. %Quantization compresses model weights, saving memory and speeding up computations. This method maintains performance while making large models more practical for resource-limited devices. It also makes fine-tuning feasible with limited computational resources. QLoRA effectively balances model complexity and performance for modern NLP tasks.

%\subsection{In-context Learning}
%LLM alignment ~\cite{wang2023aligning} is another approach aimed at achieving better results. It involves aligning large language models with human-generated text. This alignment process occurs across different stages, including data collection, training methodologies, and model evaluation. Recent advancements in alignment techniques, such as Lima ~\cite{zhou2023lima}, have shown promise by fine-tuning a 65B-parameter LlaMA model on a set of 1,000 demonstrations using LLM alignment, resulting in more accurate text generation. 
%Zero-shot prompting is a technique where a pre-trained language model performs tasks without any specific preparation or examples just for that task. The model uses broad training on diverse datasets to generate responses based on the prompt. This method showcases the model's ability to generalize knowledge across different domains, making it useful for tasks with limited labeled data.

The success of LLM-based text generation heavily depends on the prompt used. Therefore, a heavy emphasis has been paid to define the best prompt, which is known as \textit{prompt engineering}. Output Automater pattern, Template pattern, Persona pattern, Alternative Approaches pattern, Question Refinement pattern, Visualization Generator pattern, and Context Manager pattern are some of the prompt patterns proposed in the literature to assist in developing prompts~\citep{white2023prompt}. However, the best pattern depends on the task for which the LLM is used.

In the context of text generation, in-context learning~\citep{dong2022survey} has emerged as a valuable technique. This approach leverages the context specified in the prompt to make predictions based on just a few training examples. It involves extracting patterns from examples provided within the context and using them to perform a given task. However, the performance of in-context learning on downstream tasks can vary widely, depending on the number and quality of the examples.

%Few-shot prompting is a key method in in-context learning, allowing language models to handle tasks with just a few examples ~\citep{dong2022survey}. By giving the model a small number of input-output pairs, it can learn the pattern and apply it to new data. This reduces the need for extensive training for each specific task, showing how well the model can understand and generalize from limited information. Few-shot prompting demonstrates the model's ability to quickly adapt to different tasks using minimal context, making it a valuable tool in NLP tasks.

%\subsection{Improving LLM performance with human feedback}
The success of the recent LLMs is partly due to the use of Reinforcement Learning from Human Feedback (RLHF)~\citep{kirk2023understanding}. These methods aim to optimize model behavior by maximizing rewards obtained through human evaluation. RLHF is an online technique that involves training two models (policy and reward) together. %It relies on a continuous feedback loop where the policy model generates outputs that are evaluated by the reward model, with human feedback used to refine this evaluation. %This interplay allows RLHF to adapt dynamically to human preferences during training.
In contrast, Direct Preference Optimization (DPO)~\citep{rafailov2023direct} is an offline technique that simplifies the training process by focusing solely on the policy model. It eliminates the need for an interactive reward model, making it computationally less demanding. Building on DPO, Contrastive Preference Optimization (CPO)~\citep{xu2024contrastive} enhances performance by guiding models to avoid translations that have the same meaning but miss context, allowing smaller models to rival larger ones with minimal changes. {Kahneman-Tversky Optimization (KTO)~\citep{ethayarajh2024kto} takes a different approach by applying ideas from prospect theory to improve alignment. As a result, KTO has been reported to align outputs better with human choices and performs as well as or better than other methods.}

% Contrastive Preference Optimization (CPO)~\citep{xu2024contrastive} is a novel training approach designed to overcome limitations of DPO. Traditional supervised fine-tuning (SFT) methods are limited by the quality of reference data, which CPO addresses by training models to avoid generating translations those have same meaning but not perfect according to the other considerations such as context. CPO effectively guides the model towards generating superior translations and rejecting lesser ones, pushing the boundaries of LLM performance in machine translation. CPO yields significant improvement while tuning only a very low percentage of parameters of the LLM ~\citep{xu2024contrastive}. This approach can lead to major improvements, allowing medium-sized language learning models to meet or even outperform larger models 

% Kahneman-Tversky Optimization (KTO)~\cite{ethayarajh2024kto} is another novel approach to model the alignment that leverages the principles of prospect theory to optimize language model generations. It diverges from traditional preference-based methods by utilizing a binary signal to determine the desirability of outputs, rather than relying on preference data which is often scarce and costly. KTO is designed to maximize the utility of generations, aligning them more closely with human decision-making biases such as loss aversion. This method has demonstrated competitive or superior performance to existing alignment methods across various model scales, making it a promising alternative for real-world applications where preference data is limited. 

\subsection{LLM Based Evaluation}
\label{sec:LLM_eval_lit}
Evaluating the performance of LLMs for downstream tasks is expensive when done manually. Current evaluation metrics for text generation such as BLEU~\citep{papineni2002bleu} and METEOR~\citep{banerjee2005meteor}, do not sufficiently capture the complexities involved in the generated text. As a solution, some research has explored the possibility of using LLMs to evaluate the quality of the generated text. 

One prominent method in LLM-based evaluation is the LLM-Eval framework~\citep{lin2023llm}, which utilizes a unified single-prompt strategy to assess open-domain dialogue systems across multiple dimensions, including content, grammar, relevance, and appropriateness. Additionally, frameworks such as Fusion-Eval ~\citep{shu2024fusion} leverage LLMs to integrate insights from various evaluators. The G-Eval framework~\citep{liu2023g} integrates a ``Chain-of-Thought" (CoT) prompting strategy to guide LLMs in producing detailed evaluation steps, enhancing interpretability and alignment with human criteria. G-Eval adopts a form-filling paradigm for assessment, enabling fine-grained scoring by weighting evaluation scores with token probabilities. %Experimental results demonstrate that G-Eval outperforms state-of-the-art evaluators in tasks such as text summarization and dialogue generation, achieving higher correlations with human judgments. 
Furthermore, LLMs are employed to evaluate the response process itself, not just the final output~\citep{saad2023ares}. %This trend reflects a growing emphasis on dynamic evaluation, where LLMs can adapt their assessment criteria based on the task context, thereby enhancing the reliability and depth of evaluations ~\citep{li2024llms}.

However,~\cite{szymanski2024limitations} mention that LLMs are susceptible to positional, knowledge and format biases. {For example, the authors of G-Eval found that GPT-4 gave higher scores to AI summaries than to human ones. This happened even when humans actually preferred the human-written versions. This suggests that LLMs might use the same internal standards for both writing and judging text. Such a bias is risky because it can lead to ``self-reinforcement". If these biased scores are used to train the LLMs, they may simply learn to favor their own style instead of following true human quality standards.}
~\citet{chen2023exploring} also caution that  distinguishing between two subpar texts may be a challenging task for LLM based evaluation techniques. %Additionally, evaluating tasks that require complex reasoning and cognitive processes poses challenges for LLMs, as they may not effectively replicate the analytical and intuitive thinking necessary for such assessments. There is often a significant misalignment between LLM outputs and expert evaluations, particularly in specialized domains, which can hinder the integration of LLMs into critical decision-making processes. 
%These limitations highlight the need for further exploration and refinement in evaluating text quality using LLMs.% In summary, while LLMs offer exciting possibilities for quality assessment, researchers must carefully consider their limitations and explore novel techniques to enhance their performance. %According to the paper, individual score method or pairwise comparison can be used for evaluation of text generation task \cite{chen2023exploring}.

\subsection{Improving Diversity}
\label{improve_diversity}
%Beam Search is a heuristic algorithm for decoding structured predictors \citep{meister2020best}. %Beam search is used for various tasks such as Neural Machine Translations \cite{meister2020best} and Speach recognition systems \cite{hannun2014deepspeechscalingendtoend}. Diverse Beam Search uses a number of Beam Search Groups instead of using a single beam search.

According to~\citet{vijayakumar2016diverse}, lack of diversity of the decoded output of a neural sequence model leads to several problems, such as repeating the same computation without a significant gain in performance. This diversity problem can be partly attributed to the Beam Search algorithm, which is an extension of greedy search~\citep{wiher2022decodingstrategiesneuraltext} and is traditionally used to produce the LLM output. { In greedy search, the most probable token at each step is selected using greedy decoding} \citep{wiher2022decodingstrategiesneuraltext}.

While many methods such as contrastive search~\citep{su2022contrastive}, Logit Suppression~\citep{chung2023increasing} and Diverse Beam Search~\citep{vijayakumar2016diverse} have been proposed, the initial solution is to vary the decoder-related hyper-parameters~\citep{huggingfacetextgeneration}:
\begin{itemize}
    \item  Temperature: Controls the randomness of token selection by scaling the probability distribution.
    \item top-k: The quantity of vocabulary tokens with the highest likelihood that should be retained for top\-k filtering.  
    \item penalty\_alpha: This value balances the confidence of the LLM and the penalty of degeneration in contrastive search decoding.
    \item no\_repeat\_ngram\_size: All ngrams of the specified size can only appear once if no\_repeat\_ngram\_size is a positive value.    
\end{itemize}

\section{Datasets}
\label{sec:datasets}
 {In order to fine-tune an LLM for the task of MWP generation, it was imperative to acquire a dataset that includes MWPs along with pertinent details such as grade and question type. {We found a few existing MWP datasets such as Math23k ~\citep{zhang2022multi}, MAWPS ~\citep{koncel2016mawps}, MathQA ~\citep{amini2019mathqa} and SVAMP \citep{patel2021nlp}. Despite these datasets lacking curriculum-related information, we created a dataset (referred to as the {Math\_Initial\_dataset} hereafter) by combining MAWPS(4139 MWPs) and SVAMP(1000 MWPs) datasets for the initial experiments. }

In addition, we created a new MWP dataset (referred to as the {MathWizard dataset} hereafter), which includes grade and question type information. {This dataset was created by following the United States of America (USA) Mathematics curriculum called ``Common Core Standards for Mathematics"\citep{CCSSmathematicsstandards}}. Table \ref{tab: Grades & Sections} shows all the sections covered in each grade under this curriculum. {First, we identified online material that contain MWPs adhering to the Common Core Standards. However, no MWPs from those online material were included in the MathWizard dataset, to avoid any possible violation of copy right. Instead, we manually modified these MWPs (i.e. entities used in the MWP, as well as numerical values) to create new MWPs. Once created, each MWP was validated against the corresponding grade and section.}  {As an example, consider the question: “Emma has 3 crayons. Tala has 2 crayons. They put all the crayons together. How many crayons are there?” In this question, both numbers are single digits. To find the answer, the student must add 3 and 2. This is a single-digit addition problem. According to the Common Core Standards for Mathematics, this type of question belongs to the “Single-Digit Addition” section. Therefore, it can be classified as a Grade 1 question. This example shows how we assign grade levels in the MathWizard dataset based on the numerals used and the operation required to solve the problem.} This process resulted in a dataset of 4k MWPs suitable for grades 1 to 6, covering all relevant sections of the curriculum. Approximately 100 questions are included per section.

Given that preference optimization requires a dataset annotated with human preference, we created an additional dataset for the following 3 selected grade and section combinations: (Grade: 1, Section: Addition), (Grade: 3, Section: Area), and (Grade: 6, Section: Fraction). Each sample in this dataset has an accepted and a rejected MWP, corresponding to a prompt. {We reviewed (thus providing the human preference) the MWPs generated by different LLMs during the initial  experimental stages for the selected grade \& section combinations and labeled them as ``accepted" or ``rejected." This process allowed us to build the preference dataset. } A sample data instance is in Table \ref{tab:Sample data in CPO dataset}. The resulting dataset has 1005 MWPs, with approximately 350 MWPs for each selected grade-section combination. The dataset was limited to these grade-section combinations since manually creating such datasets is time-consuming.

\begin{table}
    \centering
    \begin{tabular}{|>{\centering\arraybackslash}p{0.1\linewidth}|>{\raggedright\arraybackslash}p{0.9\linewidth}|} \hline 
         Grade& Section\\ \hline 
         1& Single-digit Addition, Subtraction within 20, Addition \& Subtraction(Within 20), Count, Compare numbers, Two-digit Addition and Subtraction, Measurement, Time\\ \hline 
         2& 2 - Digit Addition, 2 - Digit Subtraction, 2 - Digit Addition \& Subtraction, Numbers to 1000, 3 - Digit Addition and Subtraction, Length in Customary Units, Length in Metric Units, Money, Geometry and Fraction Concepts\\ \hline 
         3& Multiplication, Round, Addition, Subtraction, Fractions, Area, Time, Measurement, Shapes\\ \hline 
         4& Multiplication, Division, Factors, Patterns, Addition, Subtraction, Addition \& Subtraction, Fraction, Measurement, Time\\ \hline 
         5& Decimals, Multiplication \& Division, Fractions, Measurement unit conversions, Volume, Shapes\\ \hline 
         6& Ratios and Rates, Percents, Algebraic Expressions, Equations and Relationships, Area and Polygons, Surface Area and volume of solids, Operations and Fractions, Operations with Decimals, Displaying, Analysing, and Summarizing Data\\ \hline
    \end{tabular}
    \caption{Grades \& question types according to the Common Core Standards for Mathematics}
    \label{tab: Grades & Sections}
\end{table}
}

 \begin{table}
     \centering
     \begin{tabular}{|>{\raggedright\arraybackslash}p{0.3\linewidth}|>{\raggedright\arraybackslash}p{0.3\linewidth}|>{\raggedright\arraybackslash}p{0.3\linewidth}|} \hline 
          Prompt&  Chosen& Rejected\\ \hline 
          $\langle s \rangle$ [INST] Create math word problems satisfying the following requirements:
\newline
                    [ \newline
                        Grade: 1, \newline
                        Section: Single Digit Addition, \newline
                        Number of questions: 1 \newline
                    ][/INST]$\langle /s \rangle$&  A girl has 4 toy princesses. Her cousin gives her 3 more. How many toy princesses does the girl have in total?& Eva has 7 pencils. She sharpens 1 pencil. How many pencils does Eva have left?\\ \hline
     \end{tabular}
     \caption{Sample of the dataset created for preference optimization}
     \label{tab:Sample data in CPO dataset}
 \end{table}

\section{MathWiz System}\label{sec3}
\label{sec:mehtodology}
{As discussed in Section~\ref{sec:prompt_eng_lit}, the successful application of LLMs to a task depends on multiple factors, including the LLM, the prompt, inference hyper-parameters, the use of post training techniques (fine-tuning and preference optimization), as well as the nature of in-context learning. Given that the use of LLMs for MWP generation is relatively novel, we carried out an extensive set of experiments to determine the most favorable factors for the task of MWP generation. The sequence of experiments we followed is depicted in Figure~\ref{Experiment_process}.}

{Initial experiments were carried out to identify the best open-source LLM for the task of MWP generation. Then, using the selected LLM, we experimented with several prompting patterns. Once the best prompt is selected, we adjusted the hyper-parameters applicable during the inference process, in order to increase the diversity of the generated MWPs, with minimal impact on the quality of the generated MWP. Next, we resorted to post-training experiments. This involved three steps: fine-tuning the LLM with the MathWizard dataset, preference optimizing the fine-tuned LLM, and finally further fine-tuning the preference optimized LLM. We also carried out in-context learning (few-shot prompting) experiments on the fine-tuned LLM. Solvability check was carried out as the final step. If an MWP is marked as unsolvable, it is discarded, and the generator LLM is asked to generate a new MWP. The following sub-sections describe how each of these steps were implemented.} 

%\subsection{High-Level Architecture of the system}
%The flow-chart of our methodology to implement MathWiz is shown in Figure~\ref{Experiment_process}.
%First, we carried out experiments to identify the best open-source LLM for the task. Next we experimented with different prompts, and decided on the best prompt for MWP generation. Then we carried out a set of ablation studies to identify the best set of decoder hyper-parameter values that guarantee the optimal diversity across the generated questions. As the next step, we carried out post-training, i.e. fine-tuning and direct preference optimization, using the aforementioned datasets. System implementation process is shown in Figure~\ref{fig:Final_model}. %the selected LLM with the MWP datasets mentioned above.   we selected the best-performing LLM among the open-source ones. Next, we combined the user-given values for grade, section, and question count requirements with the best resulting prompt. This prompt was built with prompt quality-improving techniques such as in-context learning. Next, the modified prompt passes to our trained LLM model for the MWP generation. Then we improve the diversity of the generated questions. After that, we fine-tuned the selected LLM with the aforementioned MWP datasets using instruction tuning and used few shot prompting. After post-processing the output of LLM, we can get a list of MWPs that satisfy the user requirement.  Next, we wanted to further improve the quality of the generated MWPs. We applied Reinforcement Learning techniques such as CPO to our trained model.

 \begin{figure} [h!]
    \centering
      \includegraphics[width=0.6\textwidth]{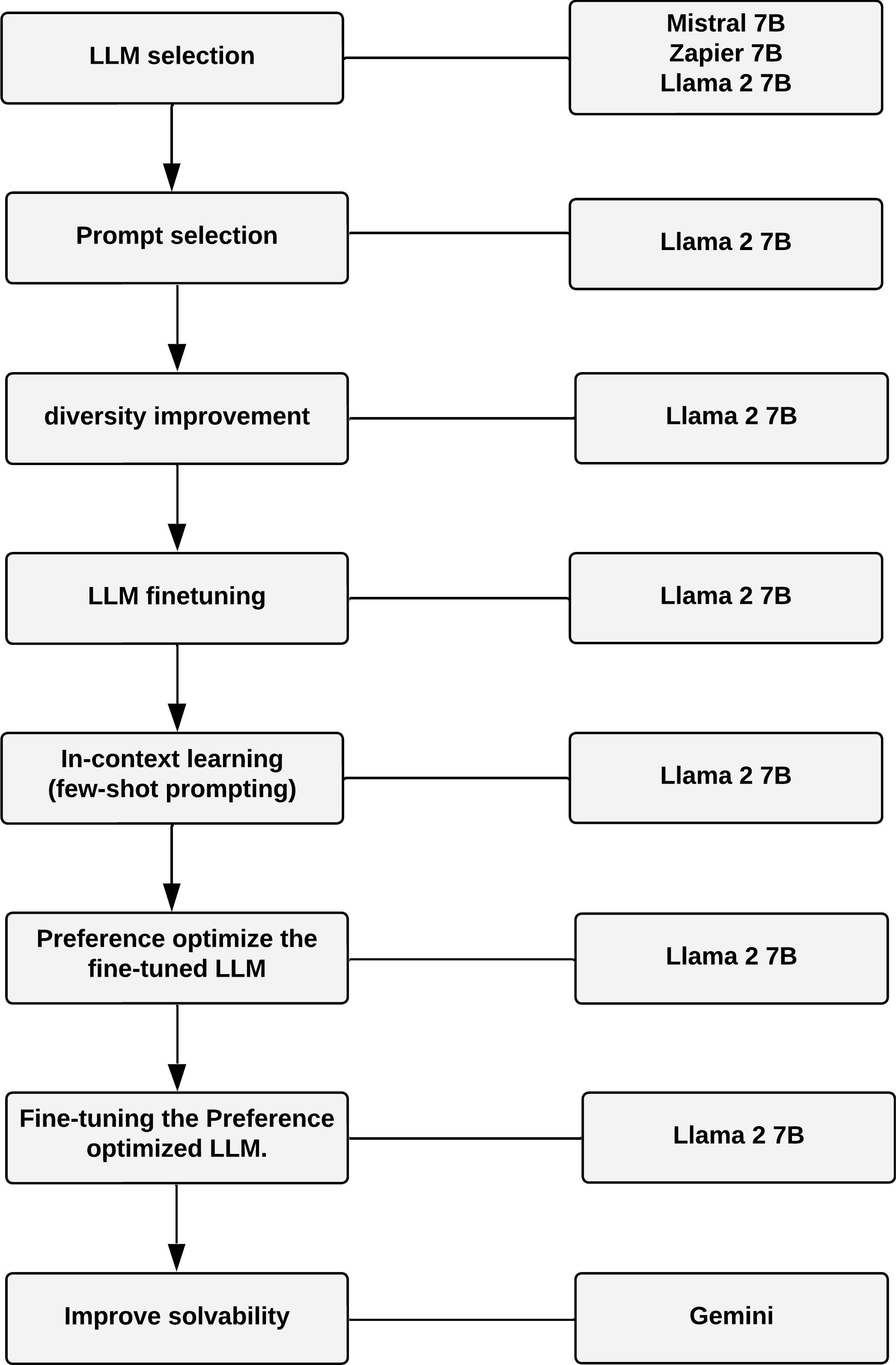}
  \caption{{Experiment process, along with the LLM used in each step}}
    \label{Experiment_process}
 \end{figure}
  \begin{figure}
      \includegraphics[width=1\textwidth]{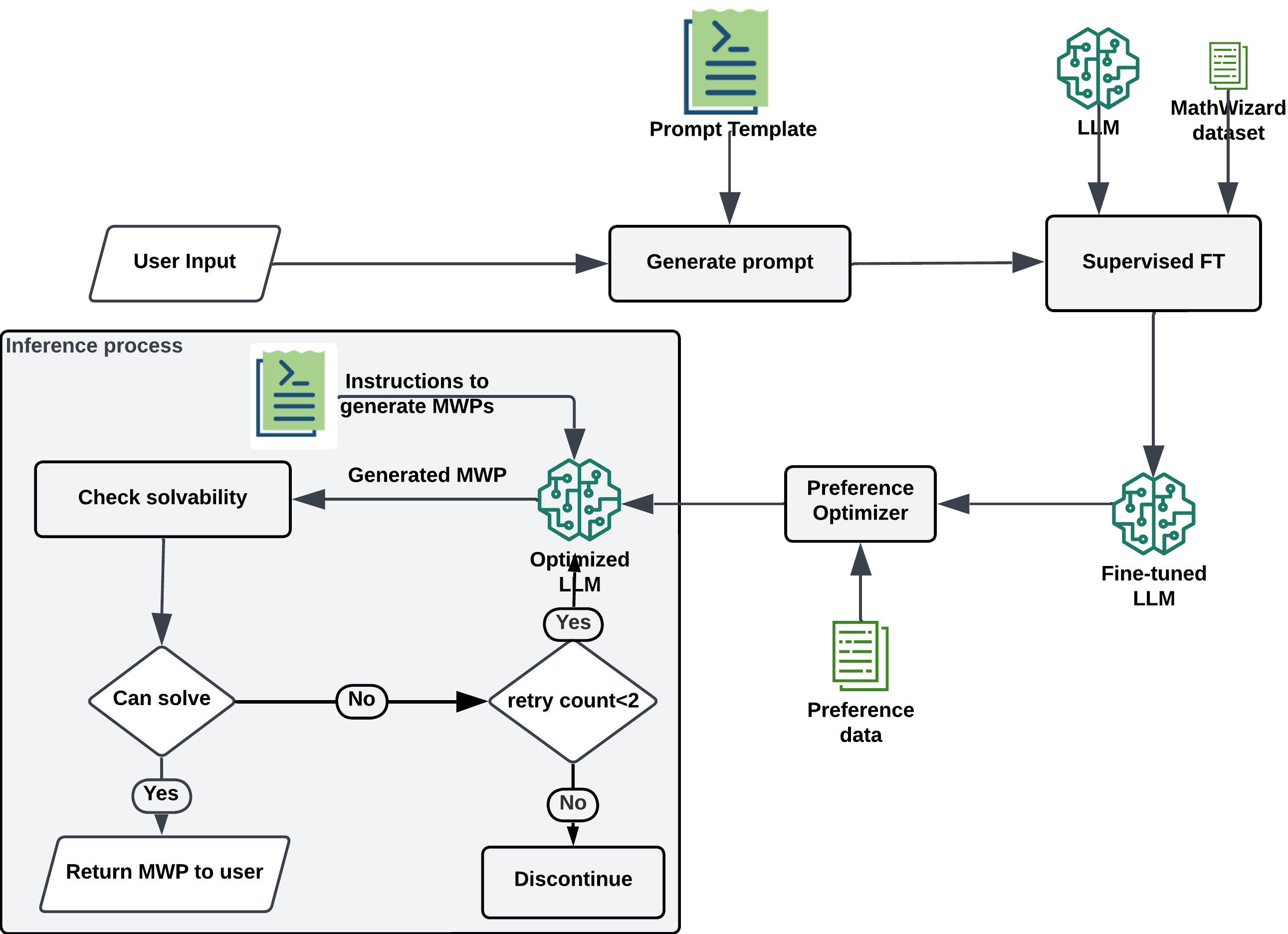}
  \caption{{Implementation of the MathWiz system, showing how a single MWP is generated from the system}}
    \label{fig:Final_model}
 \end{figure}

 %  \begin{figure}[!h] 
 % \includegraphics[width=1\textwidth]{images/meth 2.png}
 %  \caption{Further optimizing results\sr{this is not a result. anyway, i dont think this diagram is needed}}
 %    \label{Further optimizing results}
 % \end{figure}

%As further described in Section~\ref{Evaluation}, LLMs have a tendency to generate unsolvable MWPs. As a solution, during the inference stage, we incorporated a secondary LLM, to which the generated LLMs are passed. The prompt to this LLM is designed in such a way that the LLM is instructed to classify each question as solvable or unsolvable. MWPs that get marked as unsolvable are dropped, and new MWPs are generated to produce the required number of MWPs. How the solvability detection module is integrated in to the MWP generation system is shown in Fig.\ref{Resolving_Unsolvable_Problems}. 
\begin{figure}[!h]
    \centering
\fbox{\begin{minipage}[t][1cm]{0,9\textwidth}

\begin{itemize}
    \item I have 8 units. 2 of them are tens. What is this value?
    \item Levi wants to paint a picture that is 5 feet long. If he uses a brush that is 8 inches long, how many more brushes will he need?\\
\end{itemize}
\end{minipage}}
\caption{Examples for unsolvable MWPs}
\label{fig:unsolvable}
\end{figure}

  \begin{figure}[!h] 
\includegraphics[width=1\textwidth]{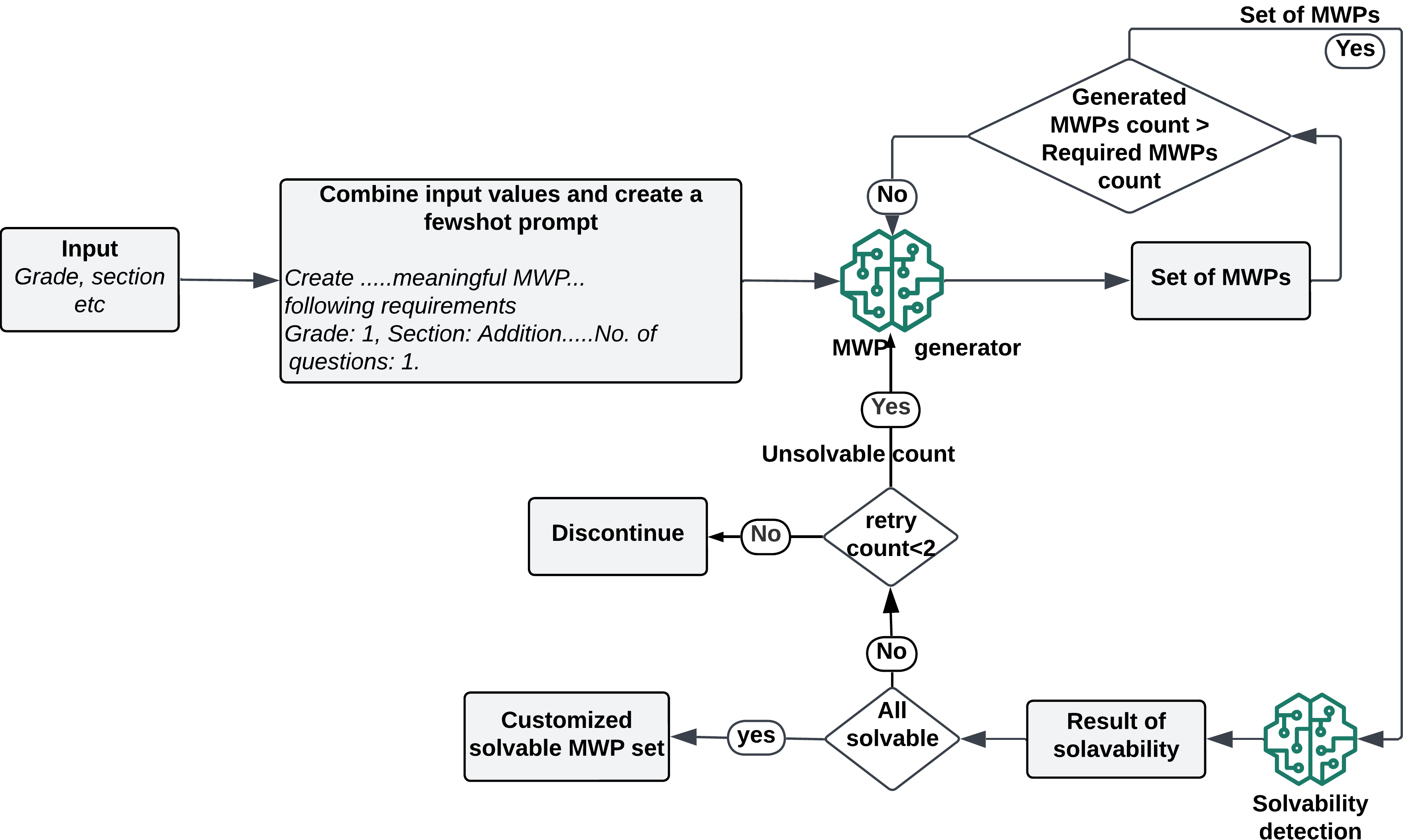}
  \caption{{How the solvability detection module is integrated with the MWP generation system when generating multiple questions during inference stage.}}
    \label{Resolving_Unsolvable_Problems}
 \end{figure}

\subsection{LLM Selection}

Based on the observations of previous studies discussed in Section~\ref{llms_for_text_generation}, we chose {Llama-2 (Llama-2-7b-chat-hf), Mistral (Mistral 7b-instruct-v0.1-hf), and Zephyr (zephyr-7b-beta)} as possible LLMs\footnote{Llama 3,4 and Deepseek were not released during the initial phase of the research}. {All the selected LLMs follow the ``Dense Transformer" architecture.} We specifically focused on open-source LLMs due to funding constraints.
 Considering the hardware limitations, we chose the 7B parameter instruct versions of these LLMs. We conducted zero-shot prompting on these LLMs to identify the best LLM for the MWP generation task. {We used the basic prompt shown in Table~\ref{tab:prompt_types} for these experiments. }

\subsection{Prompt Selection}
 {
{Considering the nature of our task, we experimented with three prompt patterns: persona, template and dialogue. The intent of the persona pattern is to give the LLM a “persona” that helps it select what types of output to generate and what details to focus on. In contrast, the template pattern instructs the LLM to produce its output in a given format. Finally, the dialogue pattern is presented as a conversation between LLM and the user. These patterns are further elaborated in Table~\ref{tab:prompt_types}, along with the basic prompt we used. As can be seen, in each of these prompts, the only input required by the user is the grade and the section corresponding to the MWP, as well as the number of MWPs that the user requires. All the prompts take the form of `instructions', which dictate the LLM to carry out the task of MWP generation. In contrast to previous research that used sequence-to-sequence Language Models such as MT5 and mBART~\citep{omegas, niyarepola2022math} that require the initial seed text or the equation of an MWP, these prompts only require meta information to generate the MWP.}

\begin{table}
    \centering
    \begin{tabular}{>{\raggedright\arraybackslash}p{1\linewidth}}
        \hypertarget{prompt1}{\textbf{Basic prompt}}\\
        Create math word problems satisfying the following requirements: \newline
\hspace*{1cm}[
\newline\hspace*{2cm}                 Grade: [GRADE], 
\newline\hspace*{2cm}                 Section: [SECTION],
\newline\hspace*{2cm}                 Number of questions: [QUESTION\_COUNT]
\newline\hspace*{1cm}        ]\\
\\

\hline
        \hypertarget{prompt2}{\textbf{Persona pattern}}\\
        Think you are a maths teacher. Create math word problems satisfying the following requirements:
\newline\hspace*{1cm}        [
\newline\hspace*{2cm}                 Grade: [GRADE],
\newline\hspace*{2cm}                 Section: [SECTION],            
\newline\hspace*{2cm}                 Number of questions: [QUESTION\_COUNT]
\newline\hspace*{1cm}        ]\\
\\
\hline
        \hypertarget{prompt3}{\textbf{Template pattern}}\\
        Think you are a maths teacher. Create math word problems satisfying the following requirements:
\newline\hspace*{1cm}        [
 \newline\hspace*{2cm}                Grade: [GRADE],
\newline\hspace*{2cm}                 Section: [SECTION],            
\newline\hspace*{2cm}                 Number of questions: [QUESTION\_COUNT]
\newline\hspace*{1cm}        ]\newline
\newline
Word in all caps is the placeholder. You should replace the placeholder with the generated problem. You should follow this template for your response: $$\langle* li \rangle PROBLEM \langle /li \rangle$$\\
\hline
        \hypertarget{prompt4}{\textbf{Dialogue pattern}}\\
        Use the following format and provide a list of math problems according to the user's requirement as output. \newline
\newline\hspace*{1cm}            User: Create math word problems satisfying the following requirements:
\newline\hspace*{2cm}                    [
\newline\hspace*{3cm}                             Grade: 1,
\newline\hspace*{3cm}                             Section: Addition,                        
\newline\hspace*{3cm}                             Number of questions: 5
\newline\hspace*{2cm}                    ]
\newline\hspace*{1cm}            Output: 1.  The school library has 15 books on the shelves. If 3 books are added to the collection, how many books does the library now have?
\newline\hspace*{2.1cm}                           2.  The classroom has 25 chairs. If 5 chairs are added to the classroom, how many chairs are now in the classroom?
\newline\hspace*{2.1cm}                           3.  The school playground has 12 swings. If 3 more swings are installed, how many swings are now on the playground?
\newline\hspace*{2.1cm}                           4.  The school has 40 students in total. If 10 more students join the school, how many students are now in the school?
\newline\hspace*{2.1cm}                           5.  The school's sports team has 30 players. If 5 more players join the team, how many players are now on the team?
\newline\newline\hspace*{1cm}            User: Create math word problems satisfying the following requirements:
\newline\hspace*{2cm}                    [
\newline\hspace*{3cm}                              Grade: [GRADE],
\newline\hspace*{3cm}                             Section: [SECTION],
\newline\hspace*{3cm}                             Number of questions: [QUESTION\_COUNT]
\newline\hspace*{2cm}                    ]
\newline\hspace*{1cm}            Output:\\
    \end{tabular}
    \caption{Example prompts created using the selected prompt types}
    \label{tab:prompt_types}
\end{table}

}

 \subsection{Improving Diversity}
 {
 The diversity in MWPs plays a crucial role in teaching Mathematics. It is important to provide questions with some variety to improve students' logical thinking. In this study, we consider two decoding strategies called Greedy Decoding and Contrastive Search. Greedy decoding generates text by selecting the most probable token at each step, which often results in deterministic but less diverse outputs. In contrast, contrastive search balances token probability and contextual degeneration penalties to produce more diverse and coherent text. As discussed in Section~\ref{improve_diversity}, there are several hyper-parameters that affect the decoding phase of an LLM.  Out of those, we experimented with \verb|top_k| and \verb|penalty_alpha|, which are associated with contrastive search~\citep{su2022contrastive}, along with \verb|no_repeat_ngram_size| and temperature~ \citep{chung2023increasing}.  We conducted a comprehensive examination of all parameter combinations (grid search), and ultimately determined the optimal values that maintain a balance between the correctness of the generated MWPs and diversity. %After obtaining the appropriate values for the parameters that enhance diversity, we proceeded with the subsequent steps.% This approach enabled us to enhance the accuracy of the generated question sets while preserving their diversity. 
 }

\subsection{Fine Tuning \& Few shot prompting}
 {
The LLM that performed the best in zero-shot prompting was selected for fine-tuning. The process of fine-tuning was carried out in the form of instruction tuning, using the best prompt out of the ones given in Table~\ref{tab:prompt_types}. {To optimize computational efficiency and mitigate high hardware requirements, we employed  QLoRA. To find the best results, we first tested different hyper-parameter settings. These included the number of training epochs, batch size, gradient norm, learning rate, and weight decay. Once we found the best combination of these settings, we kept them the same for the rest of our work.   } Initially, we carried out fine-tuning using the Math\_Initial\_dataset. Later, we fine-tuned the LLM with the MathWizard dataset that we created. 
 
{ Next, we prompted the fine-tuned LLM in a few-shots manner. This is also known as in-context learning.  Here a shot refers to an example MWP that is included in the prompt to give additional guidance to the LLM to generate new MWPs. The optimal number of shots to be used in a prompt has to be experimentally determined. On the positive side, more shots provide more information to the LLM. On the negative side, when the number of shots increases, the context length of the prompt increases, adversely affecting the reasoning capability of the LLM. Showing more examples to the LLM may also adversely affect the diversity of its output. Therefore, we experimented with 1,3, and 5 shots, to determine the optimal number of example MWPs that should be used within the prompt.}  %We tried to apply this technique to increase the accuracy of the results further. We experimented with 0, 1, 3, and 5 shots, and found that 3-shot results provided the best results which is about overall accuracy of  95.38\%. Results are shown in Table \ref{Few-shots results}. 

\subsection{Improving LLM Performance with Human Feedback}
 {As elaborated in Section~\ref{sec:evaluation}, we observed that the LLMs struggle in generating MWPs that adhere to the expected grade and question type. As a solution, we resorted to improve the performance by using human feedback. We did our initial experiments with Direct Preference Optimization (DPO), Kahneman-Tversky Optimization (KTO), and Contrastive Preference Optimization (CPO) using the preference dataset described in Section~\ref{sec:datasets}. To improve the accuracy of this optimized LLM even further, we again fine-tuned it with the MathWizard dataset, as per the recommendations of related research~\citep{thakkar2024deep, senath2025large}.} {For all these CPO, KTO and DPO based methods, QLoRA was used to avoid the hardware barrier.}

\subsection{ Improving Solvability}
During the initial zero-shot experiments, it was noted that some MWPs generated by LLMs were not solvable. Two examples are shown in Figure~\ref{fig:unsolvable}. 

Prior work shows that LLMs have remarkable Math reasoning capabilities~\citep{luo2025wizardmathempoweringmathematicalreasoning, vidal2024evaluation}.  Therefore we used a secondary LLM to evaluate the solvability of generated MWPs. Given that we do not intend to fine-tune the solvability detector for the task, we decided to employ a large commercial LLM as the solvability detector. Consequently, we experimented with the free version of Gemini, with various prompts. Best resulting prompt} is shown in Section~\ref{solvability_prompt} in Appendix.

{As shown in Figure}~\ref{Resolving_Unsolvable_Problems}, {we integrated Gemini as the final solvability checker within the MWP generation pipeline. First, output of the MWP generator is post-processed to remove irrelevant text and isolate the problem statement. This extracted MWP is then dynamically embedded into the solvability module’s prompt. The detector LLM provides a binary classification of whether the problem is solvable without considering spelling and grammar mistakes. If the MWP is solvable, the response to the corresponding prompt is TRUE. Otherwise it marks the MWP as FALSE.  If an MWP is deemed solvable, it is added to the final list for the user. For any problems identified as unsolvable, the system records the count and prompts the generator to produce replacements. This corrective loop is executed for a maximum of two iterations, after which the final solvable MWP list is delivered to the user.}

%\newpage

\section{Experiment Setup}
Our experiments were conducted using a 16 GB T4 GPU. The fine-tuning process utilized QLoRA, and hyper-parameter tuning was performed to optimize the training process. Based on initial experiments, {the optimal hyper-parameter values were identified as follows: \texttt{learning\_rate} = $2 \times 10^{-4}$, \texttt{lora\_r} = 32, \texttt{lora\_alpha} = 16, and \texttt{lora\_dropout} = 0.1. These hyper-parameter values were applied consistently throughout the subsequent experiments after fine-tuning.}

 {

For diversity improvements, first we experimented with the default parameters. We varied parameters \texttt{top\_k\_value, penalty\_alpha} and \texttt{no\_repeat\_ngram\_size} under zero-shot setting, and measured the diversity using Self-BLEU\footnote{Self-BLEU is an average of BLEU score values measured by taking each pair of sentences from the given sentence set, as a pair of measuring and reference sentences \citep{DBLP:journals/corr/abs-1802-01886}} and Jaccard~\citep{jaccard}. We identified suitable values for these parameters using Grid Search. {The grid search explored the following parameter value ranges: \texttt{top\_k\_values} from 30 to 75 in increments of 5, \texttt{penalty\_alpha} with values 0, 0.2, 0.4, and 0.6, and \texttt{no\_repeat\_ngram\_size} with values 4 and 5.}

Then we selected the most suitable parameter set for diversity improvement. {We found that the parameter combination \texttt{top\_k\_value} = 40, \texttt{penalty\_alpha} = 0.6, \texttt{no\_repeat\_ngram\_size} = 5 as the best resulting parameter combination.} %We generated 260 questions (5 questions for each grade section combination) and evaluated the diversity and quality of the question set in each step. 

 \section{Evaluation}
 \label{sec:evaluation}
Our aim is to make MathWiz capable of generating high quality, diverse MWPs while adhering to the user requirements. Therefore, we needed well-defined evaluation metrics and a consistent evaluation methodology to evaluate the performance of different experiment setups consistently. 

\subsection{MWP Error Categories}
Previous research~\citep{omegas, niyarepola2022math} has identified common issues in automatically generated MWPs. Altogether, they have defined eight possible error types. Based on our preliminary experiments, we could identify four more error types in the LLM-generated MWPs. Details about each issue type are mentioned in Table~\ref{Identified errors in the generated MWPs}. Last four error types in this table are newly defined by us. 

{Some of the error categories in Table~\ref{Identified errors in the generated MWPs} refer to surface level linguistic errors. 
Some examples are\textit{ co-reference issues}, \textit{grammatical errors} and \textit{misspellings}. In contrast, some others, including \textit{Unsolvable}, \textit{unrealistic} and \textit{unit issues}  are Mathematics-related errors. It is evident that these different types of error categories should have different levels of importance. In order to determine the relative importance of errors, we asked two Mathematics teachers to rate the severity of each error category on a scale of 1 to 5. {Higher ratings indicate greater importance.   } These ratings were used as the weights for each category. These values are also included in Table~\ref{Identified errors in the generated MWPs}.}

{In order to incorporate these weights during the evaluation, we created a ``Weighted Average" score based on the error weights~\citep{anderson2020statistic}. This score allows us to compare experiments by considering the importance of each error type. The weighted average calculation formula is defined as follows:}

\begin{equation}
\text{Weighted Average} = \frac{\sum{(\text{Accuracy \%} \times \text{Error Weight})}}{\text{Total Weight})} \times 100
\end{equation}

\begin{table}
    \centering
    \begin{tabular}{|>{\raggedright\arraybackslash}p{0.13\linewidth}|>{\raggedright\arraybackslash}p{0.3\linewidth}|>{\raggedright\arraybackslash}p{0.45\linewidth}|>{\raggedright\arraybackslash}p{0.06\linewidth}|} \hline 
         Error Type&  Description& Examples &Weight\\ \hline 
         Co-reference issue&  Inconsistency in co-reference& \textit{White is 19 years old and Black is 7 years younger than white. How old is Kalu?}
Here the second sentence has the proper noun Kalu, instead of Black. &1\\ \hline 
Trivial problem&  MWP is trivial and easy to solve& \textit{There are 50 balls. What is the total number of balls?}
Asking about the total number of balls. But the answer is already given in the question itself. &3\\ \hline 
         Grammatical errors &  Violates grammar rules of the language & \textit{Emily ran 14 mile and walked 15 mile. How much farther did Emily walk than run?}
Here mile, should be ``miles” &1\\ \hline 
         Misspellings&  Incorrectly spelled word/s & \textit{John has 3 aplles. His friend gave him 2 more aplles. How many aplles does John have now?}
Here aplles, should be ``apples". &1\\ \hline 
         Incomplete sentence&  Incomplete sentence(s) in the MWP& \textit{Samantha had 9 apples. She 4 more apples from the tree. How many apples does Samantha have now?}
The second sentence should be corrected as ``She picked 4 more apples from the tree." &2\\ \hline 
         Unsolvable &  Not enough information or has contradictions
& \textit{Tommy has 5 apples. His friend gives him more apples. How many apples does Tommy have now?}
This problem doesn't contain information about the number of apples his friend gives to Tommy. &2\\ \hline 
         Unrealistic&  Solvable but, the solution is not realistic& \textit{Amy had 8 candies. She ate 10 candies. How many candies does Amy have left?}
The answer is -2. &1\\ \hline 
         Unit issues&  An inconsistent unit is linked to a numerical value& \textit{John traveled 20 miles on his bicycle and 2 kilograms on foot. How many total miles did he travel?}
Here kilograms should be kilometers. &1\\ \hline 
         Topic unsafety&  Content is not suitable for the student.&\textit{ A Murderer killed 5 persons last January. The murderer killed 4 persons in February. How many persons were killed at the end of the February month?}
This problem contains the content about killing people. Therefore, this is not suitable for students. &3\\ \hline 
         Not satisfying grade requirement&  Not suitable for the grade& Grade: 1
Problem: \textit{Sarah spent \$ 12.50 on a book and \$ 
8.75 on a notebook. How much money did she spend in total?}
This problem needs knowledge about decimal addition. But it is not taught in grade 1. Therefore this is not suitable for grade 1. &2\\ \hline 
         Not satisfying section requirement&  Not suitable for the section& Section: Addition
Problem:\textit{ Tim had 7 toy cars. He gave away 3 toy cars to his friend. How many toy cars does Tim have now?}
This is a subtraction problem. Therefore, this problem does not satisfy the section requirement. &2\\ \hline 
         
         Not a word problem&  Not a word problem& 4 + 5 = ? &4\\ \hline         
    \end{tabular}
\caption{Identified errors in the generated MWPs }
\label{Identified errors in the generated MWPs}

\end{table}
%We developed an evaluation matrix by considering all the errors that can occur during math word problem generation. We check the presence of each error type in each generated math word problem and score 1(Error not present) or 0(Error present) according to that.

 \subsection{Evaluation Methodology}
 \label{sec:ecal_method}
 As mentioned in Section~\ref{sec:LLM_eval_lit}, there is no suitable metric to evaluate this type of text generation tasks. Therefore, human evaluation is the best way to evaluate this task. {First, we quantified the quality of the human evaluation by means of an Inter-rater-reliability (IRR) test. }Seven individuals participated as evaluators\footnote{Six are Computer Science and Engineering undergraduate, the other is a graduate in the same discipline. Two are female and the rest is male.}. We calculated IRR using Gwet's AC1 \citep{gwet2008computing} coefficient (where perfect agreement is indicated by a value of 1), which is robust to prevalence and marginal distribution issues. We divided the seven evaluators into two groups randomly and calculated the IRR between evaluators in each group.  They did the evaluation in 10 rounds and evaluated about 520 MWPs in each round. {To validate the evaluators' consistency with Mathematics teachers, two Mathematics teachers later joined the study. We then measured the inter-annotator agreement between the original evaluators and the teachers. We used the same Group 1 \& Group 2 evaluations along with the teachers' evaluation to calculate IRR agreement value. We used 100 MWPs for this calculation. }
 
However, human evaluation is a very time-consuming task. Due to the improvement in LLM-based evaluation schemes (LLM-Judges), we decided to explore the possibility of using LLMs as evaluators. We used the  recently introduced G-Eval method~\citep{liu2023gpteval} for this purpose. As discussed in Section~\ref{sec:LLM_eval_lit}, G-Eval uses an explicit score method and provides detailed evaluation by providing a numeric value for each evaluation category. Therefore, it is easy to understand the meaning of the evaluation. However, as mentioned in Section~\ref{sec:LLM_eval_lit}, LLM-based evaluation may not be entirely accurate. Therefore, in order to determine the evaluation quality of the G-Eval method, we carried out a correlation analysis against human evaluation on a sample of MWPs. We experimented with ChatGPT, Gemini, and Copilot as the LLMs in the G-Eval method. We selected 312 MWPs such that there are 2 MWPs from each grade and section combination  generated by Llama-2, Mistral, and Zephyr. This dataset was evaluated using a prompt based on the G-Eval prompt\footnote{The prompt is in Appendix~\ref{geval-prompt}.} on these three LLMs, following the error taxonomy described above.  % We obtained the correlation between human evaluation and G-Eval to determine the best LLM to be used as the evaluator. %The results are shown in the section \ref{Evaluation}. 
 \begin{table}[h]
\centering
\begin{tabular}{|l|c|c|}
\hline
\textbf{Category} & \textbf{Llama-2(\%)} & \textbf{Zephyr(\%)} \\
\hline
No Co-reference issue & \textbf{99.62} & \textbf{99.62} \\
No Grammatical error & \textbf{100} & 98.85 \\
No Misspellings & \textbf{100} & 99.62 \\
No Incomplete sentences & 99.62 & \textbf{100} \\
Solvable & 93.08 & \textbf{94.23} \\
Realistic & 95.38 & \textbf{96.15} \\
No Unit issues & \textbf{99.62} & 99.23 \\
A math word problem & \textbf{100} & \textbf{100} \\
Section relevance & 42.69 & \textbf{48.85} \\
Grade relevance & 52.69 & \textbf{56.15} \\
Topic safety & \textbf{100} & \textbf{100} \\
Non-trivial & \textbf{96.92} & 95.38 \\
\hline
\textbf{Weighted Average} & 89.63& \textbf{90.35}\\
\hline
\end{tabular}
\caption{LLM selection results}
\label{Model selection results}
\end{table}

%This result was compared against the human evaluation for the same set. The results are shown in Figure \ref{Correlation between AI models and human evaluation in each issue}. Note that we encountered limitations in accurately evaluating the criteria for ``Trivial Problem" and ``Not a Word Problem". \hl{These types of errors were too simple or lacked the usual structure of a word problem, making them hard for the model to detect } As a result, we excluded these categories from our evaluation with AI models.
\begin{figure}
    \includegraphics[width=0.3\linewidth]{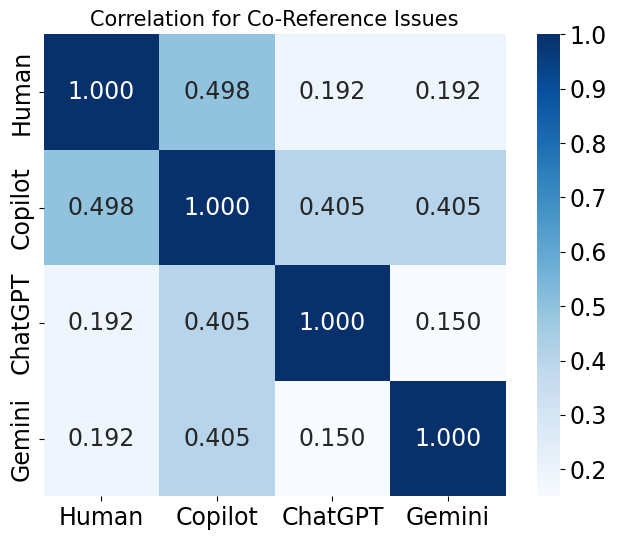}
    \includegraphics[width=0.3\linewidth]{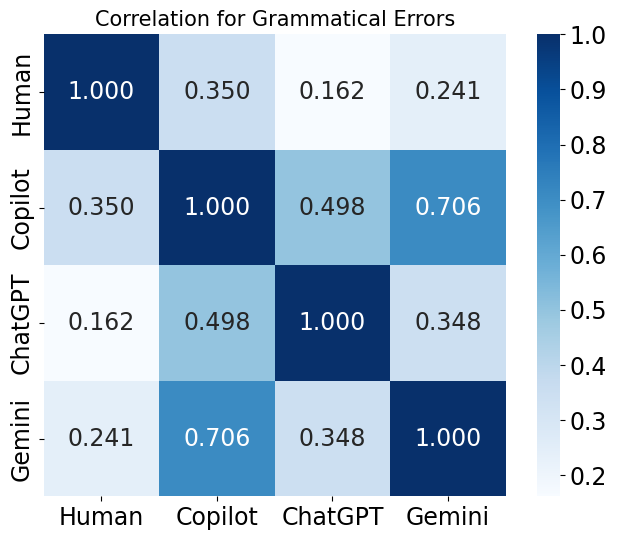}
    \includegraphics[width=0.3\linewidth]{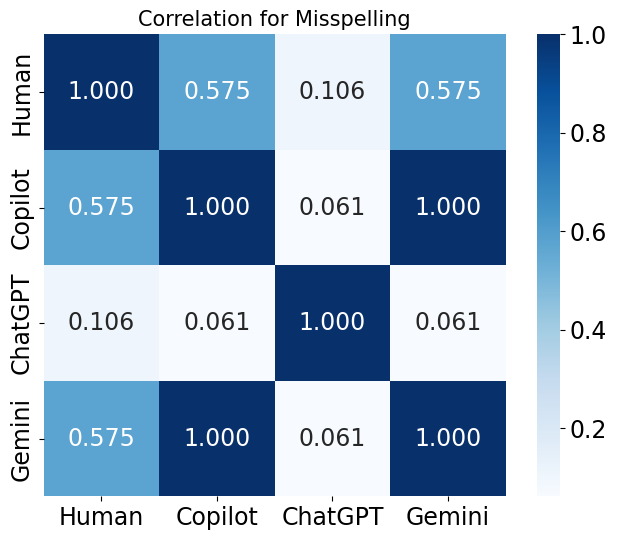}
    \includegraphics[width=0.3\linewidth]{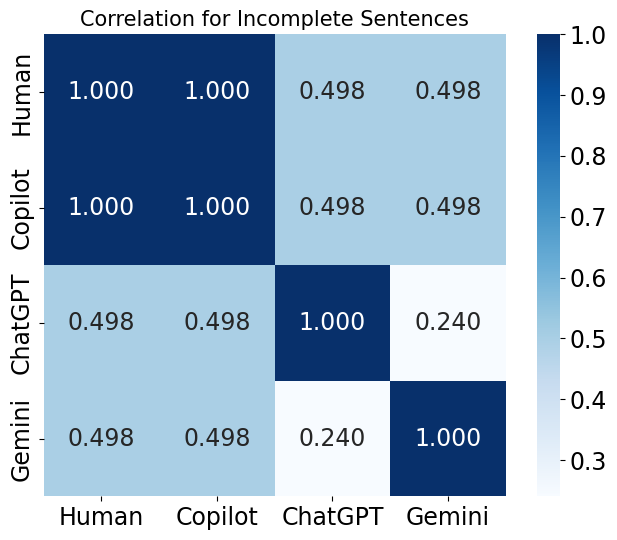}
    \includegraphics[width=0.3\linewidth]{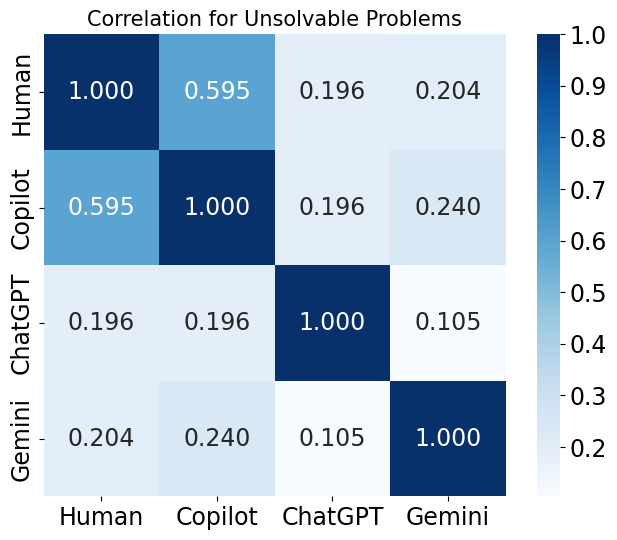}
    \includegraphics[width=0.3\linewidth]{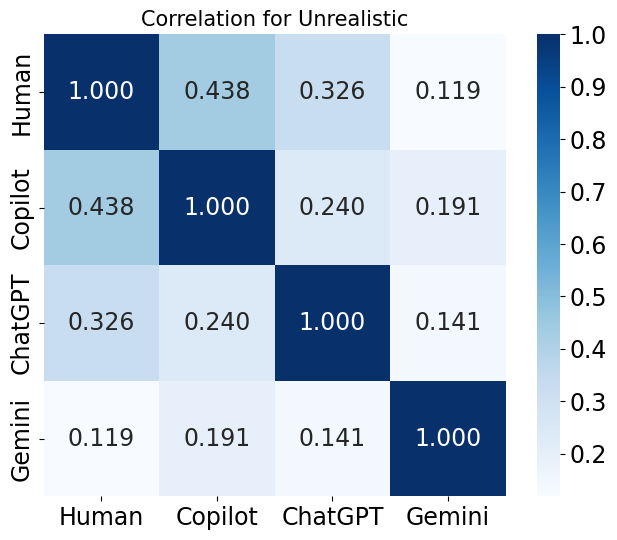}
    \includegraphics[width=0.3\linewidth]{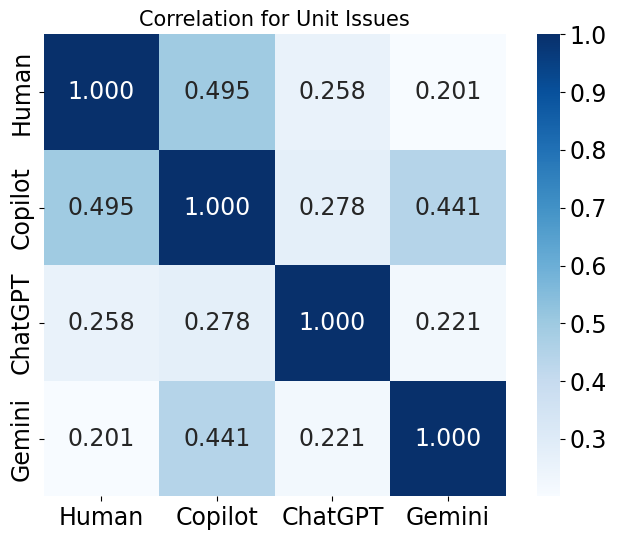}
    \includegraphics[width=0.3\linewidth]{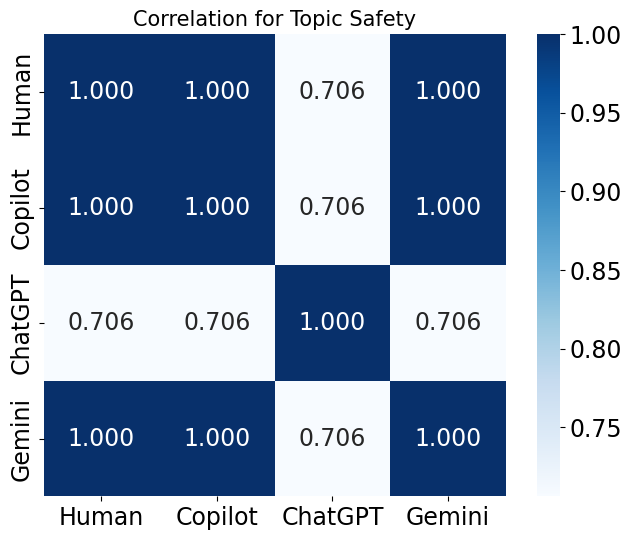}
    \includegraphics[width=0.3\linewidth]{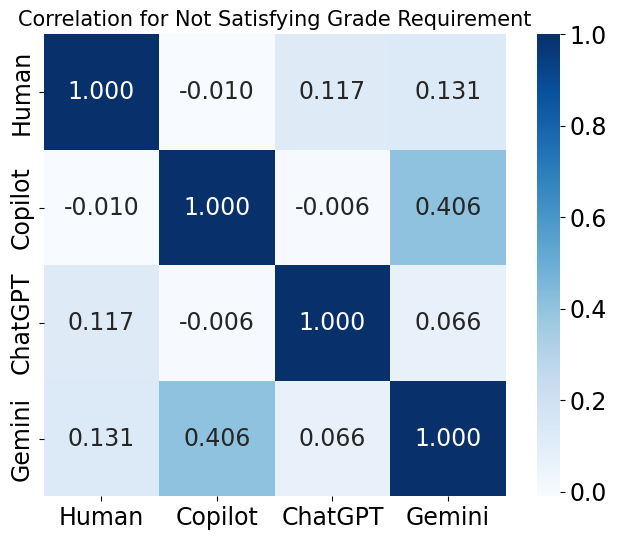}
    \includegraphics[width=0.3\linewidth]{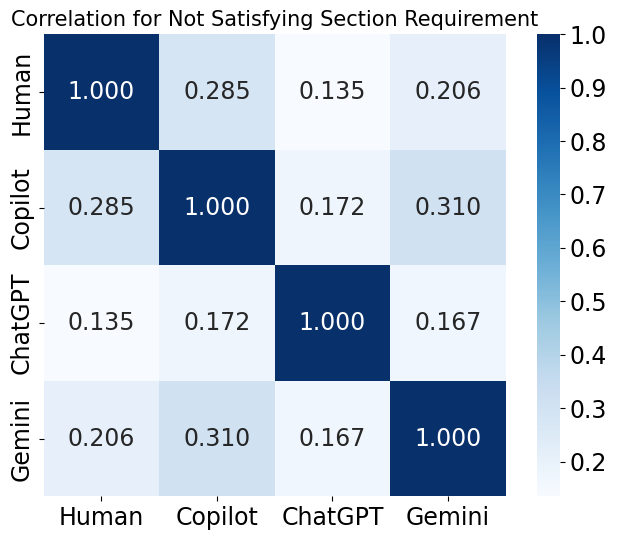}
    \caption{Correlation between AI models and human evaluation in each issue}
    \label{Correlation between AI models and human evaluation in each issue}
\end{figure}
%G-Eval conducted on Copilot showed the highest correlation against the human evaluation. Furthermore, ``Incomplete sentences" and ``Topic unsafety" are the issues for which G-Eval showed the highest correlation with human evaluation. Therefore, for the rest of the experiments, G-Eval on Copilot was used to evaluate the MWPs with respect to these two errors. Humans evaluated the MWPs for the other issues. They reviewed a total of 8,561 questions during the evaluation process, over a period of time. 

 {%In our study, we generated 260 questions covering grades 1 through 6, along with 52 math sections for each LLM we analyzed. During the question generation process, we took into account the inference time required for each question, ensuring efficiency and relevance. Then a comprehensive evaluation of the generated questions was conducted utilizing our innovative hybrid evaluation methodology.
 \iffalse
 \hl{To determine the best experiment setting at each stage, we created an "Average" score based on the error weights. This function allows us to compare experiment settings by considering the importance of each error type. The calculation for this score is defined below: 
 
 The average score is calculated by getting the weighted average of accuracy percentage of each error type by its assigned weight, then calculate the percentage. The formula is defined as follows:}

\begin{equation}
\text{Average} = \frac{\sum{(\text{Accuracy \%} \times \text{Error Weight})}}{\text{Total Weight})} \times 100
\end{equation}
\fi
\section{Results} \label{Evaluation}
\subsection{Human vs LLM Judge}
\textbf{Human Evaluation: }The results indicate a high level of agreement among the annotators, with Group 1 achieving an AC1 score of 0.9256 and Group 2 achieving 0.90427. { The evaluators also show a similar level of agreement with the teachers. Group 1 evaluators with teachers achieved an AC1 score of 0.9259, while Group 2 evaluators with teachers reached 0.89973. These results demonstrate a high level of agreement between the evaluators and the subject matter experts.}

\textbf{Performance of LLM Judge: }{Figure} \ref{Correlation between AI models and human evaluation in each issue} shows the correlation between humans and the LLM-Judges implemented on  different LLMs. Overall, Copilot produced the best result. {The results show that G-Eval is very good at finding ``Topic Safety" and ``Incomplete Sentences", which are simple structural and surface-level errors. However, it did not show a strong correlation with humans for more complex errors, such as ``Unsolvable Problems" or ``Grade/Section Requirements." This suggests that while G-Eval is reliable for checking basic language and safety, it struggles with the deep mathematical reasoning and pedagogical standards needed for MWPs. This shows that current AI evaluation still has limits when judging logical and pedagogical accuracy.  }

\subsection{Results of Experimental Steps}
{ Based on the above observations, for the rest of the experiments, G-Eval on Copilot was used to evaluate the MWPs with respect to ``Topic Safety" and ``Incomplete Sentences" errors. Humans evaluated the MWPs for the other issues. They reviewed a total of 8,561 questions during the evaluation process, over a period of time. We identified the best experiment setup considering the weighted average, as well as the number of error categories for which a given solution shows the highest result. The percentage values reported in the tables represent the proportion of MWPs that do not exhibit the corresponding error, calculated as the number of MWPs without that error divided by the total number of MWPs.}

\textbf{LLM Selection:} The results are shown in Table \ref{Model selection results}. We used zero-shot prompting with the first prompt shown in Table~\ref{tab:prompt_types}. Note that Mistral results are not shown here, because the MWPs generated by it were of extremely poor quality. While Llama-2 and Zephyr showed similar results, Zephyr's inference latency is about 10 time compared to Llama-2.  {Therefore Llama-2 was selected as the LLM for the subsequent experiments. However, we note that grade and section relevance accuracy of both LLMs is extremely low.}

\textbf{Prompt Selection:}
 { We experimented  {on Llama-2 with the zero-shot setup} with different prompt patterns listed in Table~\ref{tab:prompt_types}. Table \ref{Prompting techniques results} shows the results acquired by these different prompts. \hyperlink{prompt1}{Prompt 1} achieved the best results compared to the other prompts. Thus, this prompt was used for the future experiments. 
 \begin{table}[h]
\centering
\begin{tabular}{|l|c|c|c|c|}
\hline
\textbf{Category} & \hyperlink{prompt1}{\textbf{Prompt 1 (\%)}} & \hyperlink{prompt2}{\textbf{Prompt 2 (\%)}} & \hyperlink{prompt3}{\textbf{Prompt 3 (\%)}} & \hyperlink{prompt4}{\textbf{Prompt 4(\%)}} \\
\hline
No Co-reference issue & 99.62 & \textbf{100} & 98.83 & 98.85 \\
No Grammatical error & \textbf{100} & 98.94 & 88.72 & 99.62 \\
No Misspellings & \textbf{100} & \textbf{100} & 99.61 & 99.62 \\
No Incomplete sentences & 99.62 & \textbf{100} & 95.33 & 99.23 \\
Solvable & \textbf{93.08} & 83.45 & 91.44 & 92.69 \\
Realistic & 95.38 & 97.89 & \textbf{98.44} & 95.38 \\
No Unit issues & 99.62 & \textbf{100} & 99.22 & 97.69 \\
A math word problem & \textbf{100} & \textbf{100} & \textbf{100} & \textbf{100} \\
Section relevance & 42.69 & 41.70 & 57.71 & \textbf{59.62} \\
Grade relevance & \textbf{52.69} & 33.45 & 40.78 & 40.38 \\
Topic safety & \textbf{100} & \textbf{100} & \textbf{100} & \textbf{100} \\
Non-trivial & \textbf{96.92} & 95.77 & 94.55 & 94.62\\
\hline
\textbf{Weighted Average} & \textbf{89.63} & 87.02 & 88.65 & 89.52 \\
\hline
\end{tabular}
\caption{Results of different prompts on Llama-2 (zero-shot)}
\label{Prompting techniques results}
\end{table}
}

\textbf{Fine-tuning and few-shot prompting: }
Results of fine tuned Llama-2 is shown in Table~\ref{tab:few-shot} (under the zero-shot column).  Note that the results of Llama-2 fine-tuned with the Math\_Initial\_dataset are not shown in the results, because that experiment failed to generate MWPs with acceptable quality. We believe this is due to that dataset not having the grade level information of MWPs. The fine-tuned version of Llama-2 was tested using few-shot prompting as well. We varied the shot count as 1, 3 and 5. These results are also shown in Table~\ref{tab:few-shot}. The 5-shot evaluation achieved the best results compared to other few-shot tests.
% {We fine-tuned our model with 2 datasets that are created to improve the knowledge about the math syllabus on the Llama-2 model. The first dataset was created using the data in 2 available datasets(MWPS and SVAMP) that don't contain the details about grades on each question. The fine-tuned model with that dataset doesn't provide the MWPs in good quality. Therefore, we didn't use that dataset and fine-tuned it in future experiments. Next, our baseline model was trained with the "MathWizard" dataset. This fine-tuned model was capable of providing higher accurate results with higher improvements in section relevance and grade relevance accuracies.
 }
%\textbf{Few-shot Experiments:}
 {
 
 \begin{table}[h]
\centering
\begin{tabular}{|l|l|c|c|c|}
\hline
\textbf{Category}  &\textbf{0 shot (\%)}& \textbf{1 shot (\%)} & \textbf{3 shots(\%)} & \textbf{5 shots(\%)} \\
\hline
No Co-reference issue &\textbf{99.62}& 98.08 & 98.85 & 99.23\\ \hline 
No Grammatical error &92.69 
& \textbf{96.15} & 93.05 & 93.85 \\ \hline 
No Misspellings &85.77
& 93.85& \textbf{94.23} & 93.08 \\ \hline 
No Incomplete sentences &93.08
& 88.08 & 96.15 & \textbf{99.62} \\ \hline 
Solvable &74.23
& 73.08 & 81.85 & \textbf{81.92} \\ \hline 
Realistic &88.85
& 90 & \textbf{92.69} & 91.54 \\ \hline 
No Unit issues &97.69 & 98.85 & 96.15 & \textbf{100} \\ \hline 
Topic safety  &98.46 & \textbf{100} & \textbf{100} & \textbf{100} \\ \hline 
Grade relevance &65.38 & 66.15 & 66.93 & \textbf{79.62} \\ \hline 
Section relevance &69.23 & 75 & 75 & \textbf{81.15} \\ \hline 
Non-trivial &\textbf{92.69}& 86.15 & 89.62 & 91.92\\ \hline 
A math word problem &\textbf{100} & \textbf{100} & \textbf{100} & \textbf{100} \\
\hline
\textbf{Weighted Average}  &88.78 & 88.70 & 90.60 & \textbf{92.96} \\
\hline
\end{tabular}
\caption{ Results of few-shot prompting on fine-tuned Llama-2.}
\label{tab:few-shot}
\end{table}

}

\begin{table}
%\centering

\begin{tabular}{|>{\raggedright\arraybackslash}p{0.12\linewidth}|>{\raggedright\arraybackslash}p{0.7cm}|>{\raggedright\arraybackslash}p{0.7cm}|>{\raggedright\arraybackslash}p{0.7cm}|>{\raggedright\arraybackslash}p{0.7cm}|>{\raggedright\arraybackslash}p{0.7cm}|>{\raggedright\arraybackslash}p{0.7cm}|>{\raggedright\arraybackslash}p{0.7cm}|>{\raggedright\arraybackslash}p{0.7cm}|>{\raggedright\arraybackslash}p{0.7cm}|>{\raggedright\arraybackslash}p{0.7cm}|>{\raggedright\arraybackslash}p{0.7cm}|>{\raggedright\arraybackslash}p{0.7cm}|}\hline

%&  \multicolumn{3}{|c|}{\rotatebox{90}{Greedy Decording Strategy}}&\multicolumn{3}{|l|}{\rotatebox{90}{Combination 01 (Contrastive Search strategy)}} & \multicolumn{3}{|l|}{\rotatebox{90}{Combination 02 (Contrastive Search strategy)}} & \multicolumn{3}{|l|}{\rotatebox{90}{Combination 03 (Contrastive Search strategy)}} \\\hline
&  \multicolumn{3}{|c|}{{Combination 00}}&\multicolumn{3}{|l|}{Combination 01} & \multicolumn{3}{|l|}{Combination 02} & \multicolumn{3}{|l|}{{Combination 03}} \\\hline

 &  top\_k& pa&nrn&top\_k &pa&nrn& top\_k & pa& nrn & top\_k & pa&nrn\\\hline 
 &   N/A& N/A&0&30 & 0.4 & 4 & 40 & 0.6 & 5 & 70 & 0.2 & 4 \\\hline

No Co-reference issue (\%) &   \multicolumn{3}{|l|}{99.62}&\multicolumn{3}{|l|}{95.35} & \multicolumn{3}{|l|}{\textbf{99.62}} & \multicolumn{3}{|l|}{98.85} \\\hline

No Grammatical error (\%) &   \multicolumn{3}{|l|}{100}&\multicolumn{3}{|l|}{84.88} & \multicolumn{3}{|l|}{\textbf{97.31}} & \multicolumn{3}{|l|}{95.38} \\\hline

No Misspellings (\%) &   \multicolumn{3}{|l|}{100}&\multicolumn{3}{|l|}{82.17} & \multicolumn{3}{|l|}{\textbf{96.92}} & \multicolumn{3}{|l|}{\textbf{96.92}} \\\hline

No Incomplete sentences (\%) &   \multicolumn{3}{|l|}{99.62}&\multicolumn{3}{|l|}{92.25} & \multicolumn{3}{|l|}{\textbf{99.23}} & \multicolumn{3}{|l|}{96.15} \\\hline

Solvable (\%) &   \multicolumn{3}{|l|}{93.08}&\multicolumn{3}{|l|}{76.74} & \multicolumn{3}{|l|}{88.46} & \multicolumn{3}{|l|}{\textbf{89.62}} \\\hline

Realistic (\%) &   \multicolumn{3}{|l|}{95.38}&\multicolumn{3}{|l|}{93.80} & \multicolumn{3}{|l|}{95.00} & \multicolumn{3}{|l|}{\textbf{96.54}} \\\hline

No Unit issues (\%) &   \multicolumn{3}{|l|}{99.62}&\multicolumn{3}{|l|}{94.96} & \multicolumn{3}{|l|}{\textbf{100.00}} & \multicolumn{3}{|l|}{99.23} \\\hline

Topic safety (\%) &   \multicolumn{3}{|l|}{100}&\multicolumn{3}{|l|}{\textbf{100.00}} & \multicolumn{3}{|l|}{\textbf{100.00}} & \multicolumn{3}{|l|}{\textbf{100.00}} \\\hline

Grade relevance (\%) &   \multicolumn{3}{|l|}{42.69}&\multicolumn{3}{|l|}{\textbf{72.48}} & \multicolumn{3}{|l|}{65.00} & \multicolumn{3}{|l|}{47.88} \\\hline

Section relevance (\%) &   \multicolumn{3}{|l|}{52.69}&\multicolumn{3}{|l|}{\textbf{74.03}} & \multicolumn{3}{|l|}{54.62} & \multicolumn{3}{|l|}{55.98} \\\hline

Non-trivial (\%) &   \multicolumn{3}{|l|}{96.92}&\multicolumn{3}{|l|}{92.25} & \multicolumn{3}{|l|}{\textbf{93.46}} & \multicolumn{3}{|l|}{90} \\\hline

A math word problem (\%) &   \multicolumn{3}{|l|}{100}&\multicolumn{3}{|l|}{\textbf{99.61}} & \multicolumn{3}{|l|}{98.08} & \multicolumn{3}{|l|}{95.38} \\\hline

Weighted Average (\%)&   \multicolumn{3}{|l|}{89.63}&\multicolumn{3}{|l|}{89.45} & \multicolumn{3}{|l|}{\textbf{90.27}} & \multicolumn{3}{|l|}{87.73} \\\hline

 Self Bleu score& \multicolumn{3}{|l|}{0.021353}& \multicolumn{3}{|l|}{0.004799}& \multicolumn{3}{|l|}{\textbf{0.003976}}& \multicolumn{3}{|l|}{0.007456}\\\hline
 Jaccard score& \multicolumn{3}{|l|}{0.238161}& \multicolumn{3}{|l|}{0.187223}& \multicolumn{3}{|l|}{\textbf{0.181161}}& \multicolumn{3}{|l|}{0.193603}\\ \hline

\end{tabular}
\caption{Results for the best 3 parameter combinations. Combination 00 refers to the Greedy Decoding Strategy. Combination 01, 02, and 03 refer Contrastive Search strategy with different parameter combinations. nrn- no repeat ngram size, pa - penalty alpha.}
\label{tab:best_3_parameters}
\end{table}

\textbf{Diversity Improvement:}
Table \ref{tab:best_3_parameters} shows the results for the best 3 parameter values under the contrastive search strategy, as well the the results of the greedy decoding strategy. Figure~\ref{fig:Diversity_improvement} shows the impact of diversity improvement over {Llama-2 zero-shot result.}
According to the results for the best 3 parameters, we chose the parameter combination 02 shown in Table \ref{tab:best_3_parameters} as the best parameter combination for MWP generation. 

\begin{figure}[h]
    \centering
    \includegraphics[width=0.75\linewidth]{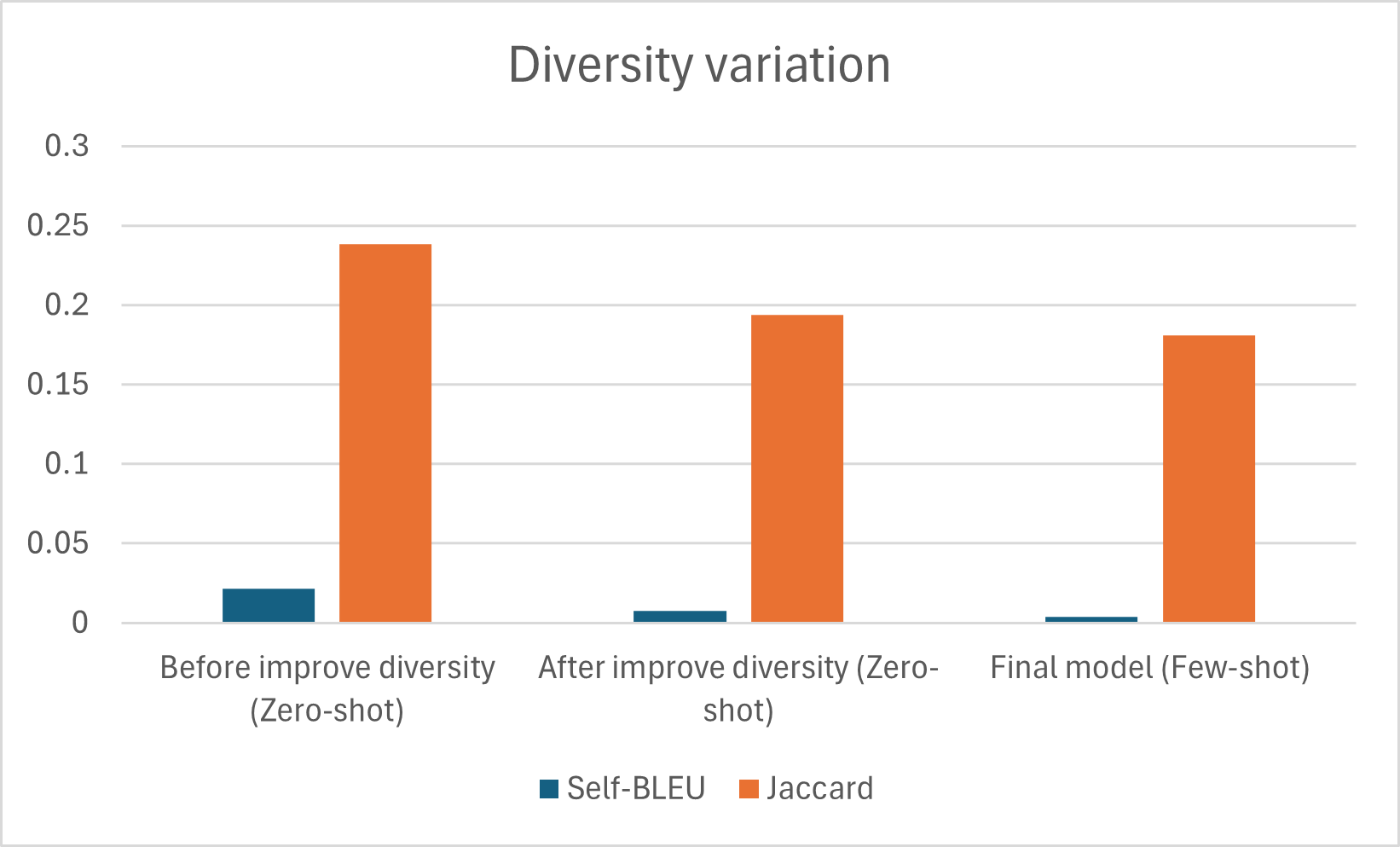}
    \caption{Diversity improvement (For both self-BLEU and Jaccard, a lower value indicates a higher diversity)}
    \label{fig:Diversity_improvement}
\end{figure}

}

\textbf{Preference Optimization:}
We experimented with KTO, DPO and CPO. DPO and KTO delivered very poor results; therefore, only the CPO results are shown in Tables~\ref{tab:CPO_results_1}, ~\ref{tab:CPO_results_2} and ~\ref{tab:CPO_results_3} in the Appendix. According to the results shown in those tables, we could achieve the highest or on-par accuracy for the categories: ``No Co-reference issue", ``No Misspellings", ``No Incomplete sentences", ``No Unit issues", ``Topic safety", ``Grade relevance", ``Section relevance", ``Non-trivial", and ``A math word problem". However, the overall accuracy declined due to issues such as an increase in the unsolvability of the generated problems.

\textbf{Solvability Module:} Table~\ref{tab:gemini_perfomance} shows the accuracy of the solvability detector, based on a manual evaluation of 200 MWPS. 
Accordingly, 88.24\% of the MWPs that are actually solvable, have been identified as solvable by the solvability detector (TPs). Similarly, 40\% of unsolvable MWPs has been identified as unsolvable (TNs). 60\% of unsolvable MWPs have been identified as solvable (FPs) and 11.76\% of solvable MWPs have been identified as unsolvable (FNs). {According to these results, it is evident that the solvability detector is not 100\% accurate in determining the solvability of an MWP. On the positive side, the solvability detector demonstrates a high accuracy in determining that an MWP is solvable, when it is indeed solvable. Therefore, whenever the solvability detector marks an MWP as solvable, we accept that decision, despite the slight margin of error. On the other hand, even though there is a possibility that the solvability detector misidentifies a solvable MWP as unsolvable, we simply discard that MWP, as a precautionary measure. }
\begin{table}[h]
  \centering
  \begin{tabular}{|>{\centering\arraybackslash}p{4cm}|>{\centering\arraybackslash}p{4cm}|}
    \hline
    Measure & Value(\%) \\ \hline
    True Positives (TP) & 88.24\\ \hline
    True Negatives (TN) & 40\\ \hline
    False Positives (FP) & 60\\ \hline
    False Negatives (FN) & 11.76\\ \hline
 Accuracy&80\\\hline
 Precision&89\\\hline
 Recall&88\\\hline
 F1&88.7\\ \hline
  \end{tabular}
  \caption{Performance of the unsolvability detection module}
  \label{tab:gemini_perfomance}
\end{table}

{The comparison of results between before and after applying the solvability detector is shown in Table \ref{tab:Result_summary} along with the results of other steps. Applying the solvability checker on top of fine-tuned Llama-2 improves the results with respect to ``unsolvable problem'', ``unrealistic problem'' and ``Trivial'' categories. This result does show that the solvability detector accomplishes its duty to a certain extent. However, there is a slight drop in the overall average accuracy. }
\iffalse
 \begin{table}[h]
\centering
\begin{tabular}{|l|l|c|}
\hline
\textbf{Category}  &\textbf{Before applying solvability technique}& \textbf{After applying solvability technique}\\
\hline
No Co-reference issue &99.23\% 
& \textbf{99.62\%}
\\ \hline 
No Grammatical error & 
\textbf{93.85\%}& 92.66\%
\\ \hline 
No Misspellings &93.08\% 
& \textbf{93.44\%}\\ \hline 
No Incomplete sentences & 
\textbf{99.62\%}& 96.14\%
\\ \hline 
Solvable &81.92\% 
& \textbf{84.17\%}\\ \hline 
Realistic & 
91.54\% 
& \textbf{94.98\%}\\ \hline 
No Unit issues &\textbf{100\%} 
& 98.84\%
\\ \hline 
Topic safety  &\textbf{100\%} 
& \textbf{100\%}
\\ \hline 
Grade relevance & 
 
\textbf{79.62\%}
& 74.52\%
\\ \hline 
Section relevance &\textbf{81.15\%}
& 78.76\%
\\ \hline 
Non-trivial & 
 
91.92\% 
& \textbf{94.59\%}
\\ \hline 
A math word problem & \textbf{100\%} 
& 99.61\%
\\
\hline
\textbf{Average}  & 
 
 \textbf{92.66\%} & 92.28\%\\
\hline
\end{tabular}
\caption{Solvability results}
\label{tab:solvability-results}
\end{table}
\fi

Note that CPO results are not shown in Table \ref{tab:Result_summary},\hl{} as that experiments were conducted only for a subset of grades and ad question types. The results in this table show that we could improve the weighted average accuracy by 3.33\% over the Llama-2 zero-shot baseline results. Most importantly, our best model beats the baseline by 26.93\%  for grade relevance and  by 38.46\% for section relevance.}

 \begin{table}[!h]
\centering
\begin{tabular}{|p{1.3in}|p{0.7in}|p{0.7in}|p{0.7in}|>{\raggedright\arraybackslash}p{0.7in}|}
\hline
\textbf{Category} & {\textbf{Llama-2 (0-shot with best prompt and diversity values) (\%)}}&  \textbf{+FTed with MathWizard data (0-shot results) (\%)}& \textbf{+FTed with MathWizard data (5-shot results) (\%)}&\textbf{{+Solvability approach applied on the 5-shot inference output (\%)}}\\
\hline
No Co-reference issue & \textbf{99.62}& \textbf{99.62}& 99.23 
 &\textbf{99.62}\\ \hline 
No Grammatical error & \textbf{97.31}& 92.69 
& 93.85
 &92.66\\ \hline 
No Misspellings & \textbf{96.92}& 85.77
& 93.08 
 &93.44\\ \hline 
No Incomplete sentences & 99.23
& 93.08 
& \textbf{99.62}&96.14\\ \hline 
Solvable & \textbf{88.46}& 74.23
& 81.92 
 &84.17\\ \hline 
Realistic & \textbf{95}& 88.85
& 91.54 
 &94.98\\ \hline 
No Unit issues & \textbf{100}
& 97.69
& \textbf{100} 
 &98.84\\ \hline 
Topic safety& \textbf{100}& 98.46& \textbf{100}&\textbf{100}\\ \hline 
Grade relevance& 65& 65.38& \textbf{79.62}&74.52\\ \hline 
Section relevance& 54.62& 69.23& \textbf{81.15}&78.76\\ \hline 
Non-trivial & 93.46& 92.69& 91.92&\textbf{94.59}\\ \hline 
A math word problem & 98.08& \textbf{100}& \textbf{100}&99.61\\
\hline
\textbf{Weighted Average} & 90.27& 88.78& \textbf{92.96}&92.56\\
\hline
\end{tabular}
\caption{Results summary (+Fted refers to the fine-tuned Llama-2 using the best prompt)}
\label{tab:Result_summary}
\end{table}

\section{Conclusion}\label{sec4}
In this paper, we presented MathWiz, an MWP generation system using the state-of-the-art LLMs. We carried out an extensive set of experiments to determine the LLM, prompt pattern, and the LLM inference parameters. We further experimented with fine-tuning and preference optimization. We achieved the best result with {5 shot inference on the fine-tuned Llama-2. Preference optimization does help in improving the result over some error categories, however it adversely affects some other categories. A larger human preference dataset that covers all error types, sections and grades is needed in the future, to fully quantify the impact of preference optimization. Similarly, the solvability detector also reduces Mathematics-related errors, however it adversely affects some errors such as grade and section relevance}. As a by-product of this research, a dataset of MWPs with human annotated errors was produced. This dataset, as well as the MathWizard dataset that we created are publicly released. 

The results reveal that LLMs find it challenging to adhere to grade and section requirements, despite the many techniques we implemented. While LLM fine-tuning and few-shot prompting alleviates this problem to a certain extent, the results are still far from optimal. As a solution, in the future, we aim to experiment with Retrieval Augmented Generation (RAG) techniques to help the LLM improve the grade and section relevance. Furthermore, we plan to investigate how MWPs can be generated according to Learning Outcomes in Mathematics curricular. We also plan to expand into multilingual MWP generation. 

\section{Declaration of generative AI and AI-assisted technologies in the manuscript preparation process}

During the preparation of this work the author(s) used ChatGPT in order to improve the language quality of the manuscript. After using this tool/service, the author(s) reviewed and edited the content as needed and take(s) full responsibility for the content of the published article.

%\newpage
\bibliographystyle{IEEEtranN}
\bibliography{sn-bibliography}% common bib file

\section{Appendix}

\subsection{Prompt of Solvability Checking}
\label{solvability_prompt}
\fbox{\begin{minipage}[t][1cm]{1\textwidth}
Try to solve the following Math Word Problem ignoring grammar mistakes and spelling mistakes. If the problem is not solvable, output ``TRUE", otherwise output ``FALSE".
\end{minipage}}

\subsection{CPO Results}
\begin{table}[!h]
\centering
\begin{tabular}{|p{0.9in}|p{0.5in}|p{0.6in}|p{0.5in}|p{0.5in}|p{0.5in}|p{0.5in}|}
\hline
 & Zero-shot on Llama-2 (\%)& Zero-shot on Llama-2 best diversity params (\%)& +FT with MathWizard dataset (\%)& {+5-shot prompting on the FTed LLM} (\%)& CPO with Llama-2 (\%)& CPO with FTed LLM (\%)\\
\hline
No Co-reference issue & \textbf{100}& \textbf{100}& \textbf{100}& \textbf{100}& \textbf{100}& \textbf{100}\\
\hline
No Grammatical error & \textbf{100}& 80& \textbf{100}& \textbf{100}& \textbf{100}& 60\\
\hline
No Misspellings & \textbf{100}& \textbf{100}& \textbf{100}& 60& \textbf{100}& 80\\
\hline
No Incomplete sentences & \textbf{100}& \textbf{100}& \textbf{100}& \textbf{100}& \textbf{100}& \textbf{100}\\
\hline
Solvable & \textbf{100}& \textbf{100}& 40& \textbf{100}& \textbf{100}& \textbf{100}\\
\hline
Realistic & \textbf{100}& \textbf{100}& 20& \textbf{100}& \textbf{100}& 80\\
\hline
No Unit issues & \textbf{100}& \textbf{100}& \textbf{100}& \textbf{100}& \textbf{100}& \textbf{100}\\
\hline
Topic safety & \textbf{100}& \textbf{100}& \textbf{100}& \textbf{100}& \textbf{100}& \textbf{100}\\
\hline
Grade relevance & \textbf{60}& 40& 0& \textbf{60}& \textbf{60}& \textbf{60}\\
\hline
Section relevance & \textbf{60}& 20& 0& \textbf{60}& \textbf{60}& \textbf{60}\\
\hline
Non-trivial & \textbf{100}& 80& \textbf{100}& \textbf{100}& \textbf{100}& \textbf{100}\\
\hline
A math word problem & \textbf{100}& \textbf{100}& \textbf{100}& \textbf{100}& \textbf{100}& \textbf{100}\\\hline
 \textbf{Weighted Average}& \textbf{93.04}& 84.38& 73.91& 91.30& \textbf{93.04}&89.57\\\hline
\end{tabular}
\caption{Results with CPO (For Grade: 1, Section: Single-digit Addition )}
\label{tab:CPO_results_1}
\end{table}

\begin{table}[!h]
\centering
\begin{tabular}{|p{0.9in}|p{0.5in}|p{0.6in}|p{0.5in}|p{0.5in}|p{0.5in}|p{0.5in}|}
\hline
 & Zero-shot on Llama-2 (\%)& Zero-shot on Llama-2 best diversity params (\%)& +FT with MathWizard dataset (\%)& {+5-shot prompting on the FTed LLM} (\%)& CPO with Llama-2 (\%)& CPO with FTed LLM (\%)\\
\hline
No Co-reference issue & \textbf{100}& \textbf{100}& \textbf{100}& \textbf{100}& \textbf{100}& \textbf{100}\\
\hline
No Grammatical error & \textbf{100}& \textbf{100}& 60& \textbf{100}& \textbf{100}& 60\\
\hline
No Misspellings & \textbf{100}& \textbf{100}& 80& \textbf{100}& \textbf{100}& \textbf{100}\\
\hline
No Incomplete sentences & \textbf{100}& \textbf{100}& \textbf{100}& \textbf{100}& \textbf{100}& \textbf{100}\\
\hline
Solvable & \textbf{100}& \textbf{100}& 80& 60& 40& 40\\
\hline
Realistic & \textbf{100}& 80& \textbf{100}& \textbf{100}& \textbf{100}& 40\\
\hline
No Unit issues & \textbf{100}& \textbf{100}& \textbf{100}& \textbf{100}& \textbf{100}& 80\\
\hline
Topic safety & \textbf{100}& \textbf{100}& \textbf{100}& \textbf{100}& \textbf{100}& \textbf{100}\\
\hline
Grade relevance & 80& \textbf{100}& 80& 60& \textbf{100}& 60\\
\hline
Section relevance & 60& \textbf{100}& 80& 60& \textbf{100}& 60\\
\hline
Non-trivial & \textbf{100}& \textbf{100}& \textbf{100}& \textbf{100}& \textbf{100}& \textbf{100}\\
\hline
A math word problem & \textbf{100}& \textbf{100}& \textbf{100}& \textbf{100}& \textbf{100}& \textbf{100}\\\hline
 \textbf{Weighted Average}& 94.78& \textbf{99.13}& 92.17& 89.57& 94.78&82.61\\\hline
\end{tabular}
\caption{Results with CPO (For Grade: 3, Section: Area)}
\label{tab:CPO_results_2}
\end{table}

\begin{table}[!h]
\centering
\begin{tabular}{|p{0.9in}|p{0.5in}|p{0.6in}|p{0.5in}|p{0.5in}|p{0.5in}|p{0.5in}|}
\hline
 & Zero-shot on Llama-2 (\%)& Zero-shot on Llama-2 best diversity params (\%)& +FT with MathWizard dataset (\%)& {+5-shot prompting on the FTed LLM} (\%)& CPO with Llama-2 (\%)& CPO with FTed LLM (\%)\\
\hline
No Co-reference issue & \textbf{100}& \textbf{100}& \textbf{100}& \textbf{100}& \textbf{100}& \textbf{100}\\
\hline
No Grammatical error & \textbf{100}& \textbf{100}& 60& \textbf{100}& 80& 80\\
\hline
No Misspellings & \textbf{100}& 80& 60& \textbf{100}& \textbf{100}& \textbf{100}\\
\hline
No Incomplete sentences & \textbf{100}& \textbf{100}& \textbf{100}& \textbf{100}& 80& \textbf{100}\\
\hline
Solvable & \textbf{100}& 80& 20& 80& 20& 40\\
\hline
Realistic & \textbf{100}& 80& \textbf{100}& 60& 80& 60\\
\hline
No Unit issues & \textbf{100}& \textbf{100}& \textbf{100}& \textbf{100}& \textbf{100}& \textbf{100}\\
\hline
Topic safety & \textbf{100}& \textbf{100}& \textbf{100}& \textbf{100}& \textbf{100}& \textbf{100}\\
\hline
Grade relevance & 40& 40& 0& 20& \textbf{60}& \textbf{60}\\
\hline
Section relevance & \textbf{60}& 40& 0& 20& \textbf{60}& \textbf{60}\\
\hline
Non-trivial & \textbf{100}& 80& 80& \textbf{100}& \textbf{100}& \textbf{100}\\
\hline
A math word problem & \textbf{100}& \textbf{100}& \textbf{100}& \textbf{100}& \textbf{100}& \textbf{100}\\\hline
 Weighted Average& \textbf{91.30}& 83.47& 69.57& 82.61& 82.61&85.22\\\hline
\end{tabular}
\caption{Results with CPO (For Grade: 6, Section: Operations and Fraction )}
\label{tab:CPO_results_3}
\end{table}

\newpage
\subsection{GEval Prompt}
\label{geval-prompt}

\begin{tcolorbox}[breakable, enhanced, width=\dimexpr\textwidth+3cm\relax,, colback=white, colframe=black, boxrule=0.5pt, arc=0pt, 
  left=5pt,
  right=5pt,
  top=10pt,
  bottom=0pt,
  boxsep=0.1pt]
Please make sure you read and understand these instructions carefully.

    You will be given one math word problem. You'll need to rate this math word problem based on the evaluation criteria given below. You have to give a rating of either 0 or 1, as explained below.\\
    \textbf{Evaluation Criteria:}-\textbf{Co-reference issues}–Give 0 if there are inconsistencies in the use of pronouns or other referring expressions within the math word problem. If there are no such errors, give 1.\\
    - \textbf{Grammatical errors}–Give 0 if the math word problem violates English grammar rules. Give 1 if there are no grammar errors.\\
    - \textbf{Misspellings}–Give 0 if the math word problem has spelling errors. Give 1 if there are no spelling errors.\\
    - \textbf{Incomplete sentences} – Give 0 if there are incomplete sentences. If all sentences are complete, give 1.\\
    - \textbf{Unsolvable problems} – Give 0 if the problem lacks sufficient information or contains contradictions. Otherwise, give 1.\\
    - \textbf{Unrealistic} – Give 0 if the solution seems unrealistic. Otherwise, give 1.\\
    - \textbf{Unit issues} – Give 0 if units are inconsistent or missing. Otherwise, give 1.\\
    - \textbf{Topic safety} – Give 0 if the content promotes violence, hatred, or is inappropriate for students. Otherwise, give 1.\\
    - \textbf{Appropriateness of grade} – Give 0 if not suitable for the specified grade. Otherwise, give 1.\\
    - \textbf{Appropriateness of question type} – Give 0 if it doesn’t match the expected question type. Otherwise, give 1.

    \textbf{Evaluation Steps:}
    \begin{enumerate}
        \item Read the user requirements.
        \item Read the math word problem.
        \item Check it against the criteria.
        \item Assign a score for each issue (0 or 1).
    \end{enumerate}

    \textbf{User requirements:}\\
    Grade: \texttt{grade} \textbar \hspace{0.3mm} Question type: \texttt{section}

    \textbf{Math word problem:}
    \texttt{{question}}

    \textbf{Misspelled words:} (List misspelled words here)

    \textbf{Solve the math word problem (step by step):}

    \textbf{Does the math word problem require calculation? (Yes/No):}

    \textbf{Evaluation Form (scores ONLY):}

    \begin{itemize}
        \item Co-reference issues
        \item Grammatical errors:
        \item Misspellings:
        \item Incomplete sentences:
        \item Unsolvable problems:
        \item Unrealistic:
        \item Unit issues:
        \item Topic safety:
        \item Appropriateness of grade:
        \item Appropriateness of question type:
    \end{itemize}
\end{tcolorbox}

\subsection{1-shot Prompt}
\label{one_shot_prompt}

\fbox{
\begin{minipage}{\textwidth}
\textbf{Instruction:} Use the following format and provide a list of math problems according to the user's requirement as output.

\vspace{0.3cm}

\textbf{User:} Create math word problems satisfying the following requirements:

\texttt{[}\\
\hspace*{1cm}\texttt{Grade: 1,}\\
\hspace*{1cm}\texttt{Section: Single-digit Addition,}\\
\hspace*{1cm}\texttt{Number of questions: 1}\\
\texttt{]}

\vspace{0.3cm}

\textbf{Output:}

1. There are 8 chairs in the reading corner and 1 table. 
How many pieces of furniture are there in total?

\vspace{0.3cm}

\textbf{User:} Create math word problems satisfying the following requirements:

\texttt{[}\\
\hspace*{1cm}\texttt{Grade: 1,}\\
\hspace*{1cm}\texttt{Section: Single-digit Addition,}\\
\hspace*{1cm}\texttt{Number of questions: 6}\\
\texttt{]}

\vspace{0.3cm}

\textbf{Generated questions:}

\#1)\#

\end{minipage}
}

\end{document}